\documentclass[journal]{IEEEtran}

\usepackage{cite}
\usepackage{amsmath,amssymb,amsfonts}
\usepackage{graphicx}
\usepackage{xcolor}
\usepackage{colortbl}
\usepackage{booktabs}
\usepackage{tikz}
\usetikzlibrary{shadows}
\definecolor{lime}{HTML}{A6CE39}
\DeclareRobustCommand{\orcidicon}{\relax%
	\begin{tikzpicture}
		\draw[lime, fill=lime] (0,0)
		circle[radius=0.16]
		node[white]{{\fontfamily{qag}\selectfont \tiny \.{I}D}};
	\end{tikzpicture}
	\hspace{-2mm}
}
\foreach \x in {A, ..., Z}{%
	\expandafter\xdef\csname orcid\x\endcsname{\noexpand\href{https://orcid.org/\csname orcidauthor\x\endcsname}{\noexpand\orcidicon}}
}

\usepackage{amsthm}
\usepackage{url}
\usepackage{multirow}
\usepackage{caption}
\usepackage{hyperref}
\usepackage[capitalise]{cleveref}

\newtheorem{lemma}{Lemma}

\definecolor{bestpink}{HTML}{FADBD8}
\newcommand{\pinkcell}[1]{\cellcolor{bestpink}#1}
\newcommand{\bestcell}[1]{\textbf{#1}}

\newcommand{\circlednum}[1]{%
	\tikz[baseline=(char.base)]{
		\node[shape=circle, fill=black, text=white, inner sep=0.5pt, font=\bfseries\sffamily\small] (char) {#1};
	}%
}

\newcommand{\premiumsquare}{%
	\tikz[baseline=-0.6ex]{
		\node[
		draw=black,
		fill=white,
		line width=1.2pt,
		drop shadow={
			shadow xshift=0.8pt,
			shadow yshift=-0.8pt,
			opacity=0.6,
			fill=gray
		},
		minimum size=5pt,
		inner sep=0pt
		] {};
	}%
}

\definecolor{vodark}{HTML}{1D1D1F}
\definecolor{vored}{HTML}{E5484D}
\definecolor{voblue}{HTML}{4F8CFF}

\begin{document}

\title{VeriOrigin: A Proactive Image Protection Framework for Provenance, Tamper Localization, and Lossless Recovery}

\author{Jiale Chen, Wei William Wang,~\IEEEmembership{Senior Member,~IEEE,}
Junnan Yin, Chongyang Shi,~\IEEEmembership{Member,~IEEE,}
\\Li Dong,~\IEEEmembership{Member,~IEEE,} Yuanman Li{}, Xiping Hu,~\IEEEmembership{Senior Member,~IEEE}
\thanks{Jiale Chen is with the Guangdong-Hong Kong-Macao Joint Laboratory for Emotional Intelligence and Pervasive Computing, Artificial Intelligence Research Institute, Shenzhen MSU-BIT University, Shenzhen 518172, China, and also with the School of Computer Science, Beijing Institute of Technology, Beijing 100081, China (E-mail: chenoly@outlook.com).}%
\thanks{Wei William Wang is with the Engineering Research Centre of Applied Technology on Machine Translation and Artificial Intelligence, Macao Polytechnic University, Macao SAR 999078, China (E-mail: weiwang@mpu.edu.mo).}%
\thanks{Junnan Yin is with the School of Cyberspace Security, Beijing Institute of Technology, Beijing, China, and also with Zhongguancun Laboratory, Beijing, China (E-mail: yinjn2023@zgclab.edu.cn).}%
\thanks{Chongyang Shi is with the School of Computer Science, Beijing Institute of Technology, Beijing 100081, China (E-mail: cy\_shi@bit.edu.cn).}%
\thanks{Li Dong is with the Department of Computer Science, Faculty of Electrical Engineering and Computer Science, Ningbo University, Ningbo 315211, China (E-mail: dongli@nbu.edu.cn).}%
\thanks{Yuanman Li is with the Guangdong Provincial Key Laboratory of Intelligent Information Processing, College of Electronics and Information Engineering, Shenzhen University, Shenzhen 518060, China (E-mail: yuanmanli@szu.edu.cn).}%
\thanks{Xiping Hu is with the School of Medical Technology, Beijing Institute of Technology, Beijing 100081, China, and also with the Guangdong-Hong Kong-Macao Joint Laboratory for Emotional Intelligence and Pervasive Computing, Artificial Intelligence Research Institute, Shenzhen MSU-BIT University, Shenzhen 518172, China (E-mail: huxp@bit.edu.cn).}%
}

\markboth{}{}

\maketitle

\begin{abstract}
Versatile watermarking, which seamlessly integrates provenance and tamper detection, has recently attracted widespread attention. In this paper, we propose \textbf{VeriOrigin}, a comprehensive and unified image protection framework that not only robustly achieves provenance and highly precise tamper localization, but also guarantees perfect recovery of the original cover in lossless channels. To realize this framework, we depart from conventional flow-based architectures. Instead, we design \textbf{XOnet}, a CNN-based invertible encoder. XOnet comprises two standard CNN-based watermarking sub-networks, Xnet and Onet, which achieve reversibility through cross residual prediction. Based on XOnet, we train the proactive protection scheme via a two-stage strategy: first optimizing the watermarking branch, followed by the tamper localization branch. During inference, the cover image undergoes sequential watermarking and tamper residual embedding. Extensive experimental results on seven datasets demonstrate that VeriOrigin achieves precise tamper localization and strong robustness, while guaranteeing lossless cover recovery for integrity authentication. Project page is available at~\url{https://veriorigin.github.io/}.
\end{abstract}


\section{Introduction}
\label{sec:intro}

\IEEEPARstart{T}{he} rapid advancement of AIGC technologies has significantly lowered the barrier to maliciously editing and redistributing images. Text-guided editing models~\cite{rombach2022high,ramesh2022hierarchical,lugmayr2022repaint,flux2024} can repaint, extend, or locally replace images based on specific masks within seconds, and the resulting tampered content is increasingly photorealistic. This severely undermines the authenticity of digital images. 

\begin{figure*}[!t]
	\centering
	\includegraphics[width=.95\textwidth]{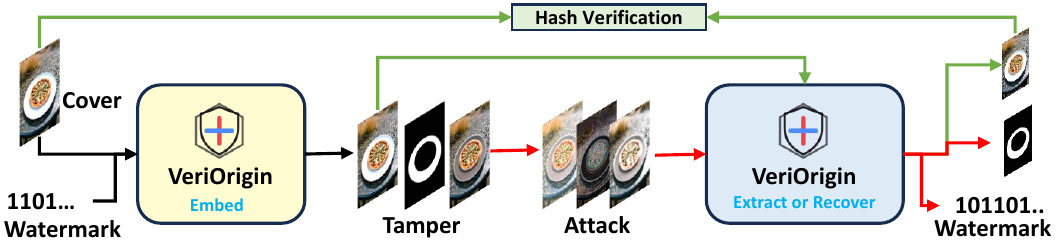}
	\caption{Overview of the VeriOrigin. After embedding via VeriOrigin, the image supports integrity authentication under lossless channels, as well as tamper localization and watermark provenance under lossy channels.}
	\label{fig:intro}
	\vspace{-10pt}
\end{figure*}

Faced with such an image, we typically need to answer two questions. \circlednum{1} \textit{where did it come from?} Answering this requires decoding the embedded information to determine its source. \circlednum{2} \textit{what modifications has it undergone since then?} Answering this requires localizing the specific tampered regions. Beyond these two, we argue that a third equally critical question is often overlooked: \circlednum{3} \textit{can we authenticate its integrity and perfectly recover the cover image when necessary?} This requires a system capable of not only verifying image provenance and localizing tampered regions, but also guaranteeing bit-exact lossless recovery to achieve integrity authentication end to end in one system.

Currently, robust watermarking primarily answers the Question \circlednum{1}. A mark is embedded so that it survives compression, resizing, and noise, while the payload identifies the source. The encoder-decoder pair is trained through a differentiable channel \cite{zhu2018hidden}. Subsequent research primarily focuses on the evolution of channel modeling: from JPEG simulators \cite{mbrs}, to more complex distortion models, and further to screen-camera captures \cite{fang2022pimog,wengrowski2019light}. These methods have demonstrated strong robustness against a wide range of complex distortions. Furthermore, a distinct category of watermarking has emerged specifically for AI-generated imagery. Instead of relying on post-processing, these methods map the watermark into latent variables, embedding the signal synchronously during the generation process to directly output a watermarked image. Specifically, existing works explore this from multiple dimensions: some root marks within the decoder weights \cite{fernandez2023the}, while others inject patterns into the initial noise \cite{wen2024tree}, and certain methods shape the latent variables for zero-cost quality embedding \cite{yang2024gaussian}. Additionally, recent studies have separately explored strategies aimed at enhancing watermark security \cite{gunn2025undetectable} or improving generation diversity \cite{yang2025t2smark}.
To address Question \circlednum{3}, robust reversible watermarking (RRW) was introduced. RRW achieves robust extraction in lossy channels and bit-exact recovery in lossless channels, thereby enabling integrity authentication \cite{chen2025crmark,chen2026drrw}. However, a fundamental gap persists across these paradigms. Existing robust watermarking is restricted to provenance, and even RRW merely extends this to recover the original cover for integrity authentication. Consequently, both can only verify \textit{whether} an image has been modified, yet neither can localize \textit{where} the tampering occurred in the first place.

To simultaneously address Questions \circlednum{1} and \circlednum{2}, versatile watermarking schemes have emerged. These methods not only embed watermarks but also precisely localize specific tampered regions after alterations occur. For instance, some works exploit the fragility of flow-based data hiding against local tampering to achieve localization \cite{zhang2024editguard}. Subsequently, learned localizers have been introduced to enhance localization robustness \cite{zhang2025omniguard}. Additionally, other studies integrate localization with watermarking within the latent space, achieving robustness against various distortions \cite{genptw2025}. Moreover, some schemes explore specific data modalities, such as face tamper detection \cite{nguyen2024laanet}, or utilize immunization strategies that enable both localization and the recovery of tampered regions \cite{ying2021imuge,ying2023imuge}. However, while these versatile methods successfully address provenance and localization, they fail to provide lossless cover recovery and integrity authentication when necessary. Therefore, exploring a proactive protection framework that seamlessly integrates provenance, localization, and lossless recovery remains a critical open challenge to date.

To address the aforementioned three challenges, we propose \textbf{VeriOrigin}. As illustrated in Fig.~\ref{fig:intro}, VeriOrigin not only achieves robust provenance but also enables precise tamper localization. Furthermore, it guarantees lossless cover recovery to achieve integrity authentication under lossless channel. To realize VeriOrigin, we first design an invertible encoder named XOnet as the foundational support. Departing from conventional flow-based architectures, XOnet adopts a CNN-based dual subnetwork design comprising Xnet and Onet. Specifically, the cover image is first partitioned into two complementary sets, X and O. Xnet predicts the pixel residuals for set O to generate the watermarked set O. Subsequently, this watermarked set O is fed into Onet to predict the residuals for set X, ultimately yielding the watermarked image. Due to the symmetric bijective nature of these cross residual predictions, the XOnet is invertible function.

Building upon the invertibility of XOnet, VeriOrigin constructs two orthogonal branches. For provenance and lossless recovery, we train a robust reversible watermark model based on XOnet, ensuring robust payload extraction under various distortions and guaranteeing bit-exact cover recovery in lossless channels. For tamper localization, we design a separate, trainable tamper sensitive residual. This residual is implemented as a content adaptive low frequency signal embedded in the complementary checkerboard half of the image. During inference, the embedding stage sequentially cascades the watermark model and the localization model. Conversely, the recovery stage reverses this process to eliminate intermediate signals layer by layer, perfectly reconstructing the pristine cover. Under lossy channels, provenance and tamper localization are directly achieved through the watermark extractor and the tamper detector, respectively. Finally, the proactive scheme is trained via a two-stage strategy: first optimizing the watermarking branch, followed by the tamper localization.

More importantly, XOnet is architecture-agnostic: any standard CNN watermark encoder can be plugged into it, since the embedding operation only relies on the encoder’s residual. This allows the Xnet and Onet components of XOnet to directly adopt off-the-shelf watermark encoders. Moreover, benefiting from the reversibility of XOnet, we further leverage it to endow existing robust watermarks with reversibility via distillation, at deployment time and without retraining.

Our \textbf{contributions} are as follows.
\begin{itemize}
	\item[\premiumsquare] We propose \textbf{VeriOrigin}, a proactive image protection framework achieving robust provenance, tamper localization, and lossless cover recovery, seamlessly integrating these tasks into a single framework.
	
\item[\premiumsquare] We design \textbf{XOnet}, a CNN-based invertible encoder that departs from conventional flow-based architectures. By employing symmetric cross residual predictions between complementary sets, XOnet guarantees strict invertibility and can host a standard CNN watermark encoder. Moreover, XOnet can distill existing robust watermarks to endow them with invertibility.
	
	\item[\premiumsquare] Extensive experiments demonstrate that VeriOrigin provides a highly competitive approach for watermark provenance and tamper localization. It not only guarantees bit-exact cover recovery in lossless channels but also keeps perceptual fidelity high.
\end{itemize}

The remainder of this paper is organized as follows. Section~\ref{sec:related} reviews the related work. Section~\ref{sec:method} details the proposed VeriOrigin framework. Section~\ref{sec:exp} provides the experimental results. Finally, Section~\ref{sec:conclusion} concludes this paper.

\section{Related Work}
\label{sec:related}

\subsection{Image Watermarking}
\label{sec:rel_wm}

Deep watermarking began by learning the residual. HiDDeN hides a message with an encoder trained
against a distortion layer \cite{zhu2018hidden}, image-in-image hiding does the same with a deeper
network \cite{baluja2017hiding}, and StegaStamp pushes the mark through a print--camera simulation so
that a printed message survives being photographed \cite{tancik2020stegastamp}; RivaGAN targets video with a
temporal attention mechanism \cite{zhang2019rivagan}, and MBRS replaces the single JPEG proxy
with a minibatch of real and simulated compression \cite{mbrs}. What varies across this
post-hoc line is the channel model, from a differentiable JPEG \cite{mbrs} to a
screen--camera capture \cite{fang2022pimog}, and the embedding itself, which can be a fully invertible transform
\cite{xu2022robust,hidden,lu2021large}.
Other axes are the two-stage training that decouples encoder and decoder \cite{liu2019a}, separability of tracing and
forgery detection \cite{wu2023sepmark}, and the fragility of the mark itself, which can be removed by
attacks \cite{zhao2025invisible,jiang2023evading}. TrustMark \cite{trustmark} and InvisMark
\cite{xu2025invismark} complete this line as general-purpose backbones for arbitrary resolutions and for provenance at scale.

The second category embeds watermarks inside the generator. Tree-Ring embeds a ring-shaped pattern into initial noise in the Fourier domain and checks watermark existence from inversely derived noise \cite{wen2024tree}. Stable Signature fine-tunes the LDM decoder to embed watermarks as latents decode into images \cite{fernandez2023the}. It embeds only a fixed bitstream and is thus a zero-bit watermark like Tree-Ring. Gaussian Shading modulates generation latents, mapping watermarks to initial latents sharing the same Gaussian distribution as sampled ones, so the watermark does not statistically harm image quality \cite{yang2024gaussian}. It obtains strong robustness from high correlation between initial noise and image content. RoSteALS embeds watermarks during generation by adding perturbations when latents become noise \cite{bui2023rosteals}. These methods bind watermarks via the generative pipeline to propagate the image, prompt and generator identity together. They aim to withstand distortions instead of detecting image edits. Recent studies examine the security of such watermarks \cite{gunn2025undetectable} and ways to maintain generated content diversity after watermarking \cite{yang2025t2smark}. Benchmarks including WAVES enable robustness evaluation \cite{an2024waves}.

Integrity authentication imposes stricter requirements than provenance verification, as it demands exact recovery of pixels after information extraction. Classical reversible watermarking schemes adopt difference expansion \cite{tian2003reversible} or histogram shifting \cite{ni2006reversible}. To pursue stronger robustness against distortions, several methods resort to invertible networks \cite{chen2025crmark,chen2026drrw}. Among these invertible models, integer flow models \cite{integerflow,van2020idf} maintain discrete operations throughout the forward process and support exact inversion. Nevertheless, these approaches are capable of realizing integrity authentication under lossless channels and robust provenance under distorted channels, while they cannot localize tampered regions.

\begin{figure}[!t]
	\centering
	\includegraphics[width=.98\linewidth]{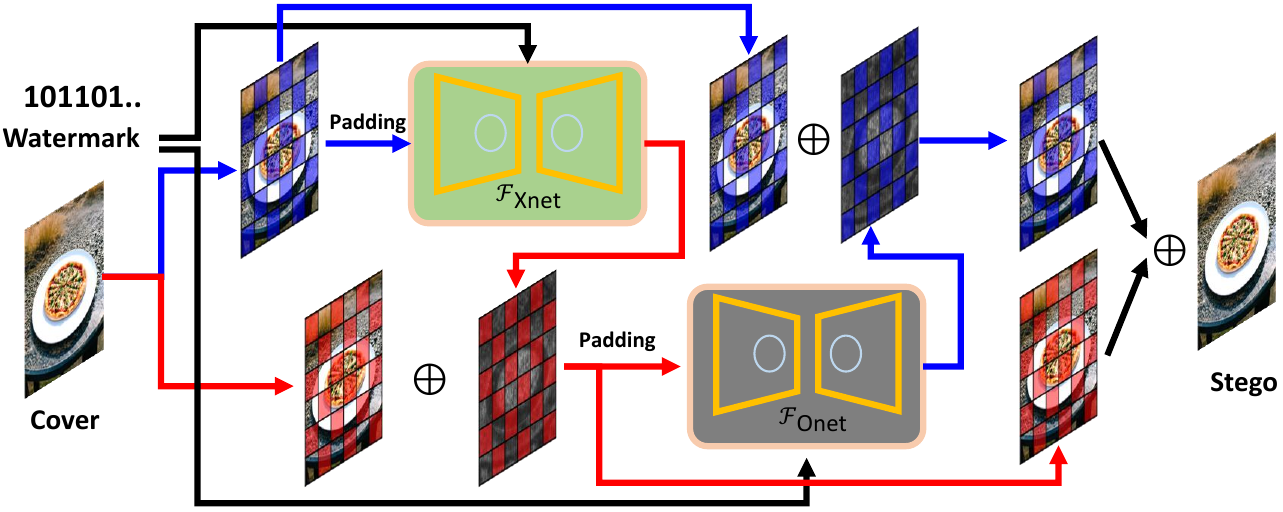}
	\caption{\textbf{XOnet, the invertible encoder of VeriOrigin.} The cover is split into two
		complementary sets, $X$ and $O$. Xnet predicts the residuals of $O$, Onet those of $X$ from the
		watermarked $O$, and running both networks backwards recovers the cover exactly.}
	\label{fig:xonet}
\vspace{-10pt}
\end{figure}

\subsection{Tamper Localization}
\label{sec:rel_tamper}

Prevalent passive forensics detects or localises known manipulation types: camera-characteristic consistency
\cite{hsu2006detecting}, noise fingerprints \cite{cozzolino2020noiseprint}, compression-artifact tracing
\cite{kwon2021cat,kwon2024learning}, copy-move localization \cite{wu2018busternet,li2018fast,islam2020doa}
and splicing localization \cite{salloum2018image}, with BusterNet separating copy-move source and target
\cite{wu2018busternet}. Universal localisers instead trace anomalies that any edit leaves behind: PSCC-Net
\cite{liu2022pscc} aggregates spatio-channel correlation progressively; TruFor \cite{guillaro2023trufor}
fuses a learned noise fingerprint with transformer features and predicts a confidence map; HiFi-Net
\cite{guo2023hierarchical} parses forgery type and attribute hierarchically; SAFL-Net \cite{sun2023safl}
forces the extractor to discard semantic content; OSN \cite{wu2022robust} targets forged images re-shared on social
networks; and ManTraNet \cite{wu2019mantra} scores local anomalies with generic filters. However, these
models learn appearance rather than provenance: without a key they cannot tell a benign edit from a
malicious one, they degrade when the channel moves, and they transfer unevenly across benchmarks. The
content is increasingly produced by generative models \cite{rombach2022high,podell2023sdxl}, with
inpainting \cite{lugmayr2022repaint} leaving little trace, which is where a proactive cue beats a passive
one.

A smaller active line makes the mark the cue. EditGuard \cite{zhang2024editguard} exploits the fragility of data hiding in flow-based models to achieve tamper localization, while training a robust watermarking model for robust provenance. OmniGuard \cite{zhang2025omniguard} introduces deep edits simulated distortions to strengthen the robustness of tamper localization. GenPTW \cite{genptw2025} protect the image in latent space, so one mark serves
provenance and localization; Furthermore, Imuge and Imuge+ represent a special class of images. By embedding protective perturbations, they endow images with the capability of self-recovery and tamper localization after manipulation\cite{ying2021imuge,ying2023imuge}. 

However, neither category supports integrity authentication: passive detectors cannot recover original pixels, and existing proactive schemes, despite enabling provenance and tamper localization, cannot achieve perfect recovery. Even if some reconstruct tampered regions, they fail to recover the original image under lossless channels. By contrast, VeriOrigin integrates these three capabilities into one proactive image protection scheme for tamper detection, provenance and integrity authentication in one unified pipeline.

\section{Method}
\label{sec:method}

\begin{figure*}[!t]
	\centering
	\includegraphics[width=\textwidth]{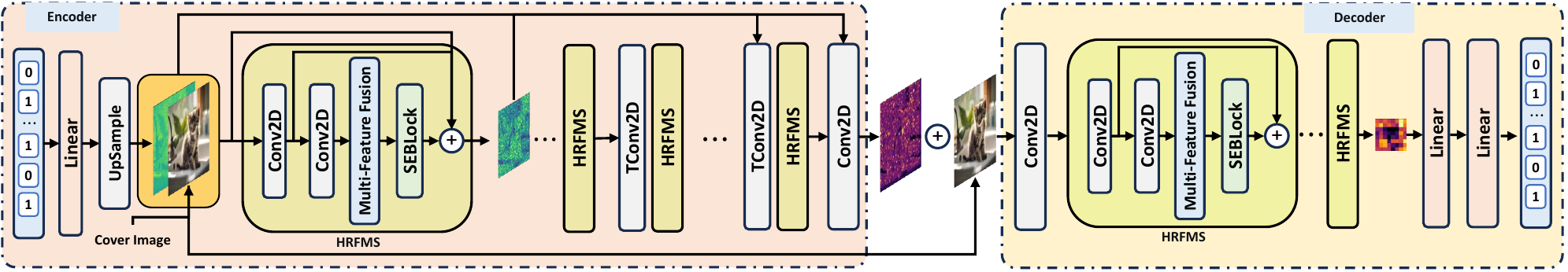}
	\caption{The watermark network architecture.}
	\label{fig:marknet}
\vspace{-10pt}
\end{figure*}

\subsection{Overview of the Invertible XOnet}
\label{sec:xonet}

The foundational support of our framework is \textbf{XOnet} (Fig.~\ref{fig:xonet}). It not only inherits the invertibility of flow-based models~\cite{realnvp,integerflow} but also retains the architectural characteristics of standard CNN watermarking networks~\cite{mbrs,hidden,trustmark,xu2025invismark}. These properties fundamentally guarantee bit-exact recovery of the original cover image, while preserving the efficiency of CNN watermarks and the priors of pretrained backbones.

\begin{figure}[!t]
	\centering
	\includegraphics[width=\linewidth]{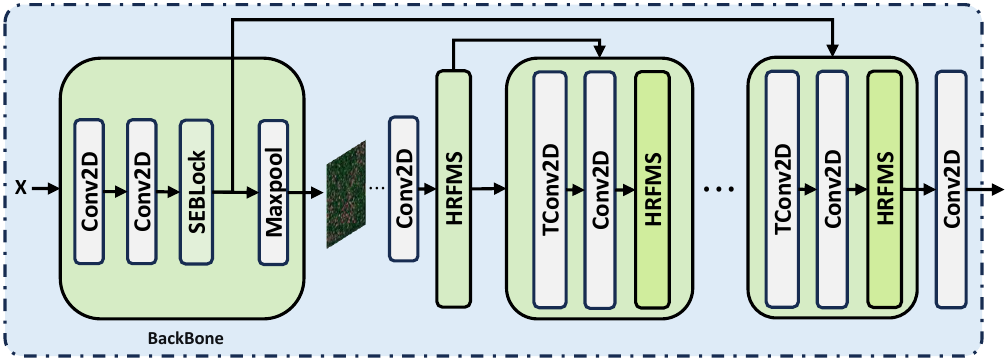}
	\caption{The tamper detection network architecture}
	\label{fig:masknet}
\vspace{-10pt}
\end{figure}

\textbf{Network Architecture.} Both branches follow a U-Net design~\cite{ronneberger2015unet}, drawn in
\Cref{fig:marknet,fig:masknet}. In watermark branch, each stage fuses the responses of convolutions with different kernel
sizes~\cite{szegedy2015going} and then applies channel and spatial attention~\cite{hu2018squeeze,woo2018cbam}; its
decoder does the same. In tamper branch, the encoder and the decoder share one structure and both apply channel and spatial
attention~\cite{hu2018squeeze,woo2018cbam}.

Let \(\mathbf{I}_{\text{org}} \in \{0, 1, \dots, 2^k\!-\!1\}^{H \times W \times C}\) denote the original cover image with \(k\)-bit depth, and \(\mathbf{w} \in \{0, 1\}^{M}\) represent the binary watermark payload. To enable invertible processing, we first partition the cover image into two complementary subsets based on a checkerboard lattice, denoted as \(X\) and \(O\). Let \(\mathbf{M}_O \in \{0,1\}^{H \times W \times C}\) denote the binary mask of set \(O\), and \(\mathbf{M}_X = \mathbf{1} - \mathbf{M}_O\) be the complementary mask for set \(X\). Then, the original cover can be expressed as:
\begin{equation}
	\mathbf{I}_{\text{org}} = \mathbf{M}_X \odot \mathbf{I}_{\text{org}}^X + \mathbf{M}_O \odot \mathbf{I}_{\text{org}}^O,
\end{equation}
where \(\odot\) denotes the \texttt{Hadamard product}. Since the masked subset \(\mathbf{I}_{\text{org}}^X\) contains zeros at all positions of set \(O\), directly feeding it into a CNN would introduce a large number of abnormal zeros, thereby affecting network training. To address this, we define a padding operator \(\mathcal{P}(\cdot)\) that fills the masked positions by interpolating from the four directly adjacent neighbors (top, bottom, left, and right), yielding the dense representation \(\hat{\mathbf{I}}_{\text{org}}^X = \mathcal{P}(\mathbf{I}_{\text{org}}^X)\).

\textbf{Forward Computation.} First, Xnet takes the padded original set \(\hat{\mathbf{I}}_{\text{org}}^X\) and the watermark \(\mathbf{w}\) as inputs to predict the pixel residuals. We apply the mask \(\mathbf{M}_O\) to restrict the prediction exclusively to the \(O\) set, followed by the rounding operator \(\mathcal{R}(\cdot)\) to generate the watermarked set \(\mathbf{I}_{\text{stego}}^O\):
\begin{equation}
	\mathbf{I}_{\text{stego}}^O = \mathcal{R}\left(\mathbf{I}_{\text{org}}^O + \mathbf{M}_O \odot \Phi_O\left(\alpha_{O} \cdot \mathcal{F}_{\text{Xnet}}(\hat{\mathbf{I}}_{\text{org}}^X, \mathbf{w})\right)\right),
\end{equation}
where \(\mathcal{F}_{\text{Xnet}}(\cdot)\) denotes the nonlinear transformation learned by Xnet, \(\alpha_{O}\) is the embedding strength, and \(\Phi_S(\mathbf{g}) = (1-\tau)\mathbf{g} + \tau\rho_S\tanh\!\big(\mathbf{g}/\sigma(\mathbf{g})\big)\), \(\tau\in\{0,1\}\), is the embedding operator of set \(S\). Here \(\sigma(\mathbf{g})\) is the standard deviation over the embedded positions of \(S\), and \(\rho_S\) is a learnable spread that keeps the residual from collapsing. The switch gives the raw embedding (\(\tau=0\)) or the normalised one (\(\tau=1\)). During task training, we employ the raw embedding; for watermark distillation, we use the normalised embedding.

\begin{figure*}[!t]
	\centering
	\includegraphics[width=.96\textwidth]{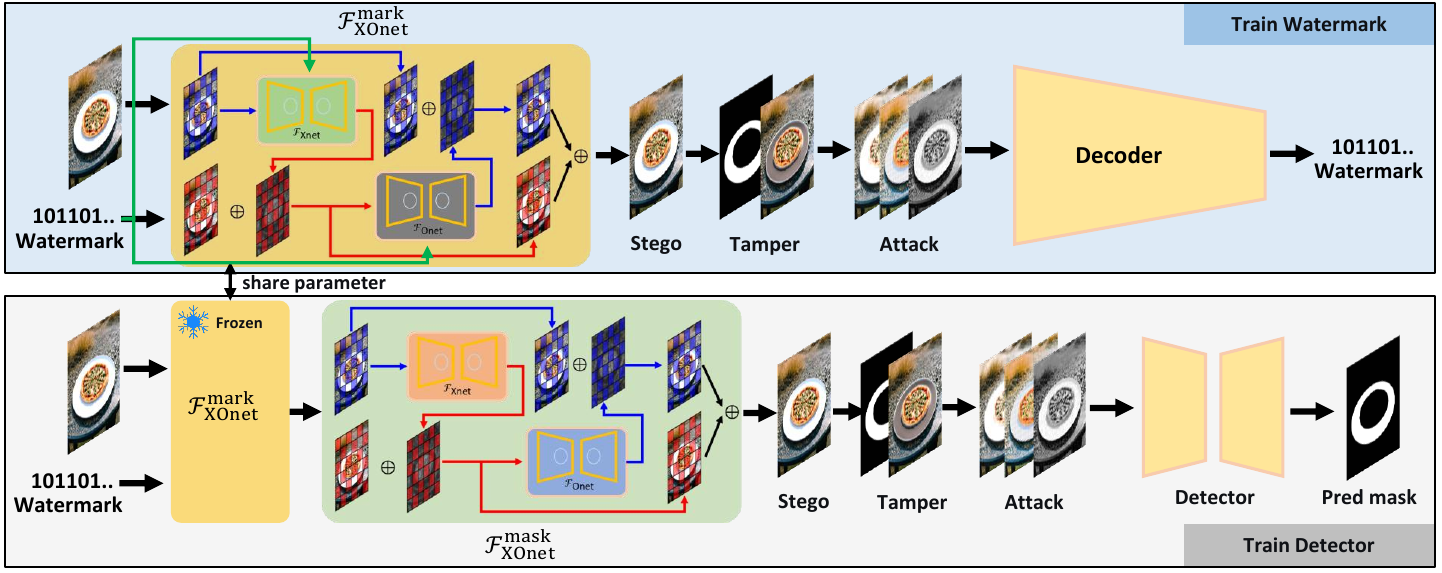}
	\caption{Two-stage training of VeriOrigin. Stage one optimizes the watermark branch through the differentiable distortion layer; stage two freezes it and trains the tamper branch on the marked image.}
	\label{fig:framework}
\vspace{-10pt}
\end{figure*}

Subsequently, this rounded set \(\mathbf{I}_{\text{stego}}^O\) is also padded into a dense representation \(\hat{\mathbf{I}}_{\text{stego}}^O = \mathcal{P}(\mathbf{I}_{\text{stego}}^O)\) and fed into Onet to predict the residuals for set \(X\):
\begin{equation}
	\mathbf{I}_{\text{stego}}^X = \mathcal{R}\left(\mathbf{I}_{\text{org}}^X + \mathbf{M}_X \odot \Phi_X\left(\alpha_{X} \cdot \mathcal{F}_{\text{Onet}}(\hat{\mathbf{I}}_{\text{stego}}^O, \mathbf{w})\right)\right).
\end{equation}
Finally, the intermediate joint image is obtained by combining the two complementary sets via their respective masks as \(\tilde{\mathbf{I}}_{\text{stego}} = \mathbf{M}_X \odot \mathbf{I}_{\text{stego}}^X + \mathbf{M}_O \odot \mathbf{I}_{\text{stego}}^O\). To ensure the validity of the stored image, we apply the clipping operator \(\mathcal{C}(\cdot)\) to obtain the valid image \(\mathbf{I}_{\text{clip}} = \mathcal{C}(\tilde{\mathbf{I}}_{\text{stego}})\). To record the exact information lost during this clipping process, we define an overflow map \(\mathbf{O}\) that captures all pixel values outside the valid range \([0, 2^k-1]\), formulated as:
\begin{equation}
	\mathbf{O} = \max(\tilde{\mathbf{I}}_{\text{stego}} - (2^k-1), \mathbf{0}) + \max(-\tilde{\mathbf{I}}_{\text{stego}}, \mathbf{0}).
\end{equation}
Subsequently, the overflow map \(\mathbf{O}\) along with the embedding strengths \(\alpha_O, \alpha_X\) and the original watermark \(\mathbf{w}\) are compressed into an auxiliary bitstream. This bitstream is embedded into \(\mathbf{I}_{\text{clip}}\) using reversible data hiding (RDH)~\cite{ni2006reversible} to obtain the final joint stego image \(\mathbf{I}_{\text{stego}}\).

\textbf{Inverse Computation.} The inverse process aims to perfectly recover the original cover image. Given the final stego image \(\mathbf{I}_{\text{stego}}\), we first apply RDH to losslessly extract the overflow map \(\mathbf{O}\), the embedding strengths \(\alpha_O, \alpha_X\), and the original watermark \(\mathbf{w}\), restoring the clipped image \(\mathbf{I}_{\text{clip}}\). 

Using an element wise indicator function \(\mathbb{I}(\cdot)\), the unbounded intermediate image is exactly reconstructed as:
\begin{equation}
	\tilde{\mathbf{I}}_{\text{stego}} = \mathbf{I}_{\text{clip}} + \mathbb{I}(\mathbf{I}_{\text{clip}} = 2^k-1) \odot \mathbf{O} - \mathbb{I}(\mathbf{I}_{\text{clip}} = 0) \odot \mathbf{O}.
\end{equation}
We then partition \(\tilde{\mathbf{I}}_{\text{stego}}\) into two complementary sets via their respective masks. Feeding the padded \(\hat{\mathbf{I}}_{\text{stego}}^O = \mathcal{P}(\tilde{\mathbf{I}}_{\text{stego}}^O)\) into Onet, the original set \(X\) is recovered by subtracting the predicted residuals:
\begin{equation}
	\mathbf{I}_{\text{org}}^X = \mathcal{R}\left(\tilde{\mathbf{I}}_{\text{stego}}^X - \mathbf{M}_X \odot \Phi_X\left(\alpha_X \cdot \mathcal{F}_{\text{Onet}}(\hat{\mathbf{I}}_{\text{stego}}^O, \mathbf{w})\right)\right).
\end{equation}

Subsequently, we pad the recovered \(\mathbf{I}_{\text{org}}^X\) into \(\hat{\mathbf{I}}_{\text{org}}^X = \mathcal{P}(\mathbf{I}_{\text{org}}^X)\). Using the known payload \(\mathbf{w}\), the original set \(O\) is exactly restored via:
\begin{equation}
	\mathbf{I}_{\text{org}}^O = \mathcal{R}\left(\tilde{\mathbf{I}}_{\text{stego}}^O - \mathbf{M}_O \odot \Phi_O\left(\alpha_O \cdot \mathcal{F}_{\text{Xnet}}(\hat{\mathbf{I}}_{\text{org}}^X, \mathbf{w})\right)\right).
\end{equation}
Ultimately, the pristine cover image is exactly reconstructed as \(\mathbf{I}_{\text{org}} = \mathbf{M}_X \odot \mathbf{I}_{\text{org}}^X + \mathbf{M}_O \odot \mathbf{I}_{\text{org}}^O\).

\textbf{Reversibility Analysis.} Mathematically, XOnet is a \textit{volume preserving} invertible additive coupling function~\cite{revnet}, defined as \(\mathcal{F}_{\text{XOnet}}: (\mathbf{I}_{\text{org}}, \mathbf{w},\alpha_O, \alpha_X) \mapsto \tilde{\mathbf{I}}_{\text{stego}}\). Its inverse function is formulated as \(\mathcal{F}_{\text{XOnet}}^{-1}: \mathbf{I}_{\text{stego}} \mapsto (\mathbf{I}_{\text{org}}, \mathbf{w},\alpha_O, \alpha_X)\). The strict invertibility relies on losslessly decoding the auxiliary bitstream (containing \(\alpha_O, \alpha_X, \mathbf{O}\), and \(\mathbf{w}\)) from \(\mathbf{I}_{\text{stego}}\). Crucially, transmitting \(\mathbf{w}\) and the strengths circumvents potential neural decoder errors that would break exact invertibility. Therefore, knowing these strengths and the overflow map precisely reconstructs the unbounded stego image. This enables XOnet to generate identical residuals, perfectly recovering the original cover. As long as this auxiliary information stays within the RDH capacity, the pristine image is exactly recovered.

\vspace{-10pt}
\subsection{Overview of the Training of VeriOrigin}
\label{sec:framework}

As illustrated in Fig.~\ref{fig:framework}, the training of VeriOrigin is divided into two stages. In the first stage, we train a robust reversible watermark based on XOnet. In the second stage, we train the tamper detection model. The detailed training procedures and their losses are described below.

\textbf{Training Watermark.} We define the watermark encoder as \(\mathcal{F}_{\text{XOnet}}^{\text{mark}}\) and the watermark extractor as \(\mathcal{E}_{\text{mark}}\). Let \(\mathcal{N}(\cdot)\) denote the differentiable distortion layer. The data inference pipeline follows a standard end to end process as \((\mathbf{I}_{\text{org}}, \mathbf{w}) \xrightarrow{\mathcal{F}_{\text{XOnet}}^{\text{mark}}} \tilde{\mathbf{I}}_{\text{stego}} \xrightarrow{\mathcal{N}} \mathbf{I}_{\text{noised}} \xrightarrow{\mathcal{E}_{\text{mark}}} \hat{\mathbf{w}}\). 

A significant advantage of this design is that both \(\mathcal{F}_{\text{XOnet}}^{\text{mark}}\) and \(\mathcal{E}_{\text{mark}}\) are CNN-based models, making the training process identical to standard CNN watermarking. This flexibility is not available in flow-based models. The total loss function for this stage comprises four components and is defined as:
\begin{equation}
	\begin{aligned}
		\mathcal{L}_{\text{mark}} &= \lambda_{\text{w}} \mathcal{L}_{\text{BCE}}(\hat{\mathbf{w}}, \mathbf{w}) + \lambda_{\text{mse}} \mathcal{L}_{\text{MSE}}(\tilde{\mathbf{I}}_{\text{stego}}, \mathbf{I}_{\text{org}}) \\
		&\quad + \lambda_{\text{lpips}} \mathcal{L}_{\text{LPIPS}}(\tilde{\mathbf{I}}_{\text{stego}}, \mathbf{I}_{\text{org}}) + \lambda_{\text{p}} \mathcal{L}_{\text{pen}}(e)(\tilde{\mathbf{I}}_{\text{stego}}),
	\end{aligned}
\end{equation}
Here, \(\mathcal{L}_{\text{BCE}}\) is the binary cross entropy loss for watermark extraction, while \(\mathcal{L}_{\text{MSE}}\) and \(\mathcal{L}_{\text{LPIPS}}\) enforce pixel level fidelity and perceptual quality. The coefficients \(\lambda\) balance the contribution of each term. Both \(\mathcal{F}_{\text{XOnet}}^{\text{mark}}\) and \(\mathcal{E}_{\text{mark}}\) follow the feature-operator design of standard CNN watermarks.

\textbf{Training Tamper Detection.} The second stage keeps the watermark branch frozen and loads its pretrained encoder \(\mathcal{F}_{\text{XOnet}}^{\text{mark}}\) into the pipeline, so that the detector is trained on the same residual the watermark layer embeds instead of on a clean cover. This is what makes the localiser adaptive to the watermark: it learns the statistics of the marked image rather than treating them as noise. Let \(\mathcal{F}_{\text{XOnet}}^{\text{det}}\) denote the tamper detection encoder, \(\mathcal{T}\) the tampering function, \(\mathcal{N}(\cdot)\) the differentiable distortion layer, and \(\mathcal{D}\) the detection network. The complete inference pipeline for this stage is explicitly formulated as: \(\mathbf{I}_{\text{org}} \xrightarrow{\mathcal{F}_{\text{XOnet}}^{\text{mark}}} \tilde{\mathbf{I}}_{\text{stego}} \xrightarrow{\mathcal{F}_{\text{XOnet}}^{\text{det}}} \mathbf{I}_{\text{prot}} \xrightarrow{\mathcal{T}} \mathbf{I}_{\text{tampered}} \xrightarrow{\mathcal{N}} \mathbf{I}_{\text{noised}} \xrightarrow{\mathcal{D}(\mathbf{I}_{\text{noised}})} \mathbf{M}_{\text{pred}}\).

Concretely, the watermark encoder \(\mathcal{F}_{\text{XOnet}}^{\text{mark}}\) first generates the stego image \(\tilde{\mathbf{I}}_{\text{stego}}\) from the original cover \(\mathbf{I}_{\text{org}}\), and the tamper detection encoder \(\mathcal{F}_{\text{XOnet}}^{\text{det}}\) then embeds the tamper sensitive residual in it to produce the protected image \(\mathbf{I}_{\text{prot}}\). The tampering function \(\mathcal{T}\) alters \(\mathbf{I}_{\text{prot}}\) into \(\mathbf{I}_{\text{tampered}}\), which \(\mathcal{N}\) degrades into \(\mathbf{I}_{\text{noised}}\). The detector reads \(\mathbf{I}_{\text{noised}}\) alone; the inverse \((\mathcal{F}_{\text{XOnet}}^{\text{det}})^{-1}\) is kept for the recovery path of Section~\ref{sec:xonet}, which restores the cover exactly at the decoder.

To jointly optimize protection and localization, the detection stage keeps the fidelity and penalty terms; only the BCE term differs: it is computed between the predicted tamper mask and the ground truth mask, \(\lambda_{\text{w}} \mathcal{L}_{\text{BCE}}(\mathbf{M}_{\text{pred}}, \mathbf{M}_{\text{gt}})\), rather than between extracted and embedded message bits. Both stages apply the same overflow penalty, which remains essential to each of them~\cite{chen2025crmark}. Because the watermark branch is frozen, this stage updates only the tamper encoder and detector.

\definecolor{bestblue}{HTML}{A8C8E8}
\newcommand{\wmcell}[1]{\cellcolor{bestblue}#1}
\setlength{\textfloatsep}{5pt plus 2pt minus 2pt}
\renewcommand{\topfraction}{0.98}
\renewcommand{\bottomfraction}{0.9}
\renewcommand{\textfraction}{0.03}
\renewcommand{\floatpagefraction}{0.85}
\setcounter{topnumber}{4}
\setcounter{bottomnumber}{2}
\setcounter{totalnumber}{8}

\subsubsection{Dynamic Weighting}\label{sec:s_loss}

 The penalty weight is not fixed in either branch. Both branches follow the same schedule:
\begin{equation}
	\lambda_{\text{pen}}(e)=
	\begin{cases}
		\lambda_{\text{pen}}^{0}, & e \le e_0,\\[2pt]
		\left\lfloor\dfrac{e-e_0+\Delta}{\Delta}\right\rfloor \lambda_{\text{lateral}}, & e > e_0,
	\end{cases}
	\label{eq:s_ramp}
\end{equation}
with \(e_0 = 2000\), \(\Delta = 25\) epochs, and \(\lambda_{\text{lateral}} = 1\). This initial delay \(e_0\) is critical: increasing the penalty weight prematurely could disrupt the early learning of watermark embedding and tamper detection. The effect of this ramp is that each branch first learns the optimal spatial positions to embed the residual, and is only subsequently pushed to minimize the magnitude at those locations, rather than uniformly shrinking the residual across the entire image.

\subsection{Extension to High-Bit Depths}\label{sec:s_hbd}

Reversible watermarking is critical for legal and scientific evidence: \(12\)/\(16\)-bit satellite scenes are reanalyzed years later under military or legal custody, and clinical studies are reread and archived with annotations~\cite{coatrieux2009}; irreversible perturbation corrupts records, while reversible data hiding restores covers bit for bit after extraction~\cite{tian2003,ni2006,shi2016}. Yet 8-bit networks trained on \([0,255]\) clip or quantize native-depth buffers, destroying exactness before embedding. \textbf{How can an 8-bit pretrained network handle \(12\)/\(16\)-bit imagery bit-exactly?} Retraining would forfeit 8-bit priors. We propose the Lossless Invertible Mapping (LIM) for high-bit-depth images.

\subsubsection{Lossless Invertible Mapping}\label{sec:s_map}
The network operates at 8-bit precision, while the high-bit-depth information is carried entirely by the residual. For a native image \(\mathbf{I}_b\) with \(b \in \{12, 16\}\), let
\(m_b = \min(\mathbf{I}_b)\) and \(M_b = \max(\mathbf{I}_b)\) be its extremes, and let
\[
s_b = \max\left\{1, \left\lceil (M_b - m_b) / 255 \right\rceil\right\}
\]
be the integer scale that spreads that range over the 8-bit domain. The mapping proceeds in four steps:
(i) Remove the offset and quantize to 8 bit: \(\mathbf{u}_8 = \lfloor (\mathbf{I}_b - m_b) / s_b \rfloor\), which lies in \([0,255]\) because \(\mathbf{I}_b - m_b \le M_b - m_b \le 255 s_b\).
(ii) Apply the reversible encoder \(\mathcal{F}_{\text{XONet}}\) with watermark \(\mathbf{w}\):
\[
\mathbf{s}_8=\mathcal{F}_{\text{XONet}}(\mathbf{u}_8,\mathbf{w})=\mathbf{u}_8+\mathbf{r}_8,
\]
where \(\mathbf{r}_8\) is an integer residual, and
\[
\mathcal{F}_{\text{XONet}}^{-1}\big(\mathcal{F}_{\text{XONet}}(\mathbf{u}_8,\mathbf{w}),\mathbf{w}\big)=\mathbf{u}_8.
\]
(iii) Map the residual back to native units as \(s_b \mathbf{r}_8\), add an integer shift \(t\) in native units, and form
\[
\mathbf{I}^{\text{stego}}_b = \mathbf{I}_b + s_b \mathbf{r}_8 + t.
\]
The shift \(t\) is recorded and removed at the receiver; it introduces only a small constant offset but eliminates most downward overflow, since the residual is predominantly negative on dark content.
(iv) Clip every sample of \(\mathbf{I}^{\text{stego}}_b\) that leaves \([0, 2^b - 1]\) and store its unclipped value in a side stream, embedded with \(\mathbf{w}\), \(m_b\), \(M_b\) and \(t\) through RDH~\cite{ni2006}; the receiver restores these samples and inverts the mapping.

\begin{lemma}[Lossless \(b\)-bit mapping]
	\label{lem:lossless_mapping}
	Let \(b \in \{12,16\}\), \(m_b=\min(\mathbf{I}_b)\), \(M_b=\max(\mathbf{I}_b)\) and
	\(s_b=\max\{1,\lceil (M_b-m_b)/255\rceil\}\), so that
	\(\mathbf{u}_8=\lfloor(\mathbf{I}_b-m_b)/s_b\rfloor\in[0,255]\) for every sample of \(\mathbf{I}_b\), and let
	\(\mathcal{F}_{\text{XONet}}(\cdot,\mathbf{w})\) be a bijection on the 8-bit domain whose inverse
	\(\mathcal{F}_{\text{XONet}}^{-1}(\cdot,\mathbf{w})\) satisfies
	\[
	\mathcal{F}_{\text{XONet}}^{-1}\big(\mathcal{F}_{\text{XONet}}(\mathbf{u}_8,\mathbf{w}),\mathbf{w}\big)=\mathbf{u}_8.
	\]
	Then for any integer shift \(t\), Steps (i)--(iii) are invertible, and every sample is recovered as
	\(\mathbf{I}_b=\mathbf{I}^{\text{stego}}_b-s_b \mathbf{r}_8 - t\), with
	\(\mathbf{r}_8=\mathcal{F}_{\text{XONet}}(\mathbf{u}_8,\mathbf{w})-\mathbf{u}_8\).
\end{lemma}

\begin{proof}
	Encoding forms
	\[
	\mathbf{u}_8=\left\lfloor \frac{\mathbf{I}_b-m_b}{s_b} \right\rfloor,\quad
	\mathbf{s}_8=\mathcal{F}_{\text{XONet}}(\mathbf{u}_8,\mathbf{w})=\mathbf{u}_8+\mathbf{r}_8,
	\]
	and produces \(\mathbf{I}^{\text{stego}}_b=\mathbf{I}_b+s_b\mathbf{r}_8+t\). 
	Write \(\mathbf{I}_b-m_b=s_b\mathbf{u}_8+e\) with \(0\le e<s_b\). Subtracting \(t\) and mapping back to 8 bit with the floor operator, using \(\lfloor x+n\rfloor=\lfloor x\rfloor+n\) for integer \(n\), gives:
	\[
	\left\lfloor \frac{\mathbf{I}^{\text{stego}}_b-t-m_b}{s_b} \right\rfloor
	=\left\lfloor \frac{\mathbf{I}_b-m_b}{s_b}+\mathbf{r}_8 \right\rfloor
	=\mathbf{u}_8+\mathbf{r}_8
	=\mathbf{s}_8.
	\]
	Since \(\mathcal{F}_{\text{XONet}}^{-1}\) is exact, \(\mathbf{u}_8=\mathcal{F}_{\text{XONet}}^{-1}(\mathbf{s}_8,\mathbf{w})\) and \(\mathbf{r}_8=\mathbf{s}_8-\mathbf{u}_8\), so \(\mathbf{I}_b=\mathbf{I}^{\text{stego}}_b-s_b\mathbf{r}_8-t\) is exact in integer arithmetic.
\end{proof}

\textbf{Reversibility analysis.} \Cref{lem:lossless_mapping} establishes exact invertibility of Steps (i)--(iii), and Step (iv) is lossless, so the stored image is recovered exactly. Operationally, the receiver extracts \(\mathbf{w}\), \(m_b\), \(M_b\) and \(t\), restores the clipped samples, subtracts the shift \(t\) from the stego image, derives \(s_b = \max\{1,\lceil (M_b - m_b)/255\rceil\}\), maps it back to the 8-bit domain with \(\lfloor(\mathbf{I}^{\text{stego}}_b - t - m_b)/s_b\rfloor\), applies \(\mathcal{F}_{\text{XONet}}^{-1}(\cdot,\mathbf{w})\) to recover \(\mathbf{u}_8\) and \(\mathbf{r}_8\), and subtracts \(s_b \mathbf{r}_8 + t\) exactly.

\begin{figure*}[ht!]
    \centering
    \includegraphics[width=\textwidth]{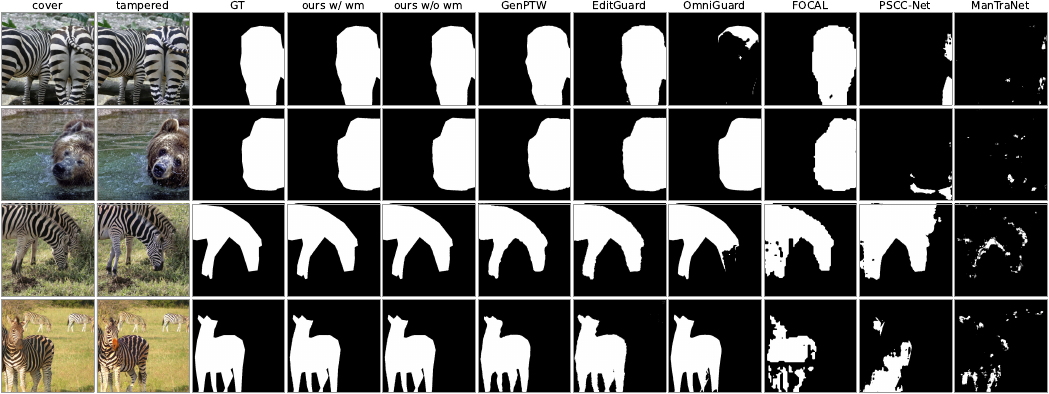}
    \caption{Localization on four TGIF2 samples. Columns: the cover, the tampered image, the ground truth, ours with ``w/ wm'', ours ``w/o wm'', GenPTW \cite{genptw2025}, EditGuard \cite{zhang2024editguard}, OmniGuard \cite{zhang2025omniguard}, FOCAL \cite{focal2023}, PSCC-Net \cite{liu2022pscc} and ManTraNet \cite{wu2019mantra}.}
    \label{fig:exp_merged}
\end{figure*}

\section{Experiments}
\subsection{Setup and Evaluation Protocol}
\label{sec:exp}

\textbf{Datasets.} We evaluate on three public manipulation datasets, the 512-pair CocoGlide dataset built by GLIDE~\cite{glide2022} inpainting on COCO~\cite{lin2014microsoft}, the 300-pair COVERAGE dataset~\cite{wen2016coveragea}, and CASIA v2.0~\cite{dong2013casia}, of which we use 1{,}828 spliced pairs with masks over eight categories; we further test on TGIF2~\cite{mareen2026tgif2}, selecting 500 pairs produced by five of its six inpainting models (FLUX.1 [schnell], [dev] and Fill [dev]~\cite{flux2024}, Stable Diffusion 2~\cite{rombach2022high} and Photoshop), 100 each, ensuring a thorough and balanced evaluation across these diverse generative models. For document forgery we use the whole testing split of AIForge-Doc~\cite{aiforgedoc}, \(613\) triples of a pristine page, an AI-forged page and the field mask, and for native-depth imagery three LIDC-IDRI CT series~\cite{armato2011lidc}, \(400\) slices in total, and two Sentinel-2 L2A acquisitions~\cite{sentinel2}, reported as \(400\) CT samples and \(360\) chips; all seven sets are scored by the same metric code, so the numbers are comparable across corpora. 

\textbf{Baselines.} For tamper detection we compare against the active localization methods EditGuard~\cite{zhang2024editguard}, GenPTW~\cite{genptw2025} and OmniGuard~\cite{zhang2025omniguard} and the passive methods PSCC-Net~\cite{liu2022pscc}, ManTraNet~\cite{wu2019mantra} and FOCAL~\cite{focal2023}; for watermark provenance we compare against TrustMark~\cite{trustmark}, InvisMark~\cite{xu2025invismark} and CRMark~\cite{chen2025crmark}, each reproduced from its official implementation and rigorously run through the exact same distortion and metric code for an unbiased evaluation. 

\textbf{Metrics.} Visual quality is measured by PSNR, SSIM and LPIPS, tamper localization by F1 and IoU, and watermark extraction by bit accuracy, over seven conditions for localization and eighteen distortions for the payload.

\subsubsection{Evaluation Protocol and Metrics}\label{sec:s_proto}
We adopt standard evaluation metrics and define them once. For a cover \(\mathbf{I}\) and a delivered image \(\hat{\mathbf{I}}\), the fidelity is measured by \(\mathrm{PSNR} = 10\log_{10}((2^k-1)^2/\mathrm{MSE}(\hat{\mathbf{I}},\mathbf{I}))\) for a \(k\) bit image, and LPIPS~\cite{zhang2018lpips}. For a message \(\mathbf{w}\) and its estimate \(\hat{\mathbf{w}}\), payload accuracy is the bit accuracy. For a predicted mask \(\hat{\mathbf{M}}\) and the ground-truth mask \(\mathbf{M}\), we report:
\begin{equation}
	\mathrm{F1} = 2|\hat{\mathbf{M}}\cap\mathbf{M}|/(|\hat{\mathbf{M}}|+|\mathbf{M}|),
	\mathrm{IoU} = |\hat{\mathbf{M}}\cap\mathbf{M}|/(|\hat{\mathbf{M}}\cup\mathbf{M}|),
	\nonumber
\end{equation}
and the pixel accuracy \(|\{\hat{\mathbf{M}}=\mathbf{M}\}|/(H W)\). The predicted mask is binarized at a threshold of \(0.5\).

\subsection{Training Details}
\label{sec:s_stage}

\textbf{Stage I: Training Watermarking.} The watermark branch is trained with the standard end-to-end training strategy. The cover and watermark bits enter the XONet, and the coupling is trained end to end against a differentiable distortion layer that draws one attack per batch from Gaussian blur, median filtering, Gaussian noise, salt-and-pepper, a JPEG surrogate, resizing, cropping, cropout, hue, brightness and saturation, and applies it to the whole batch. The reversible mapping parameter is set to \(\tau = 0\), so the residual is learned directly without applying a perceptual prior. The decoder uses a straight through estimator~\cite{bengio2013estimating} for the rounding function \(\mathcal{R}(\cdot)\), keeping its gradient alive through quantization. We optimize with AdamW (\(\beta_1 = 0.5\), \(\beta_2 = 0.999\), \(\epsilon = 10^{-6}\), weight decay \(10^{-5}\)) at a constant learning rate of \(10^{-4}\), training exclusively at a fixed \(512 \times 512\) resolution.\(\lambda_{\text{mse}} = \lambda_{\text{lpips}} = 1.0\), \(\lambda_{\text{secret}} = 0.5\), \(\lambda_{\text{pen}} = 1.0\) and \(\tau=0\). Covers are drawn from DIV2K~\cite{div2k} and the photographic training split, resized to \(512 \times 512\) and randomly cropped. The watermark length is 100 bit payload, the same budget every other method carries.

\textbf{Stage II: Training Tamper Detection.} We train the tamper detection branch with the same end-to-end strategy and keep the watermark branch frozen. Its encoder generates \(512\times512\) stego images; we crop them into \(256\times256\) patches and feed the patches to the tamper encoder, so the detector observes the residual. Patch training focuses the detector on local residual patterns rather than global content, enhancing generalization capability on smooth regions. At inference we apply the detector to VeriOrigin's \(512\) output. We train the detector on triples of cover, tampered image and mask using five attacks: blur, median filter, JPEG quality \(50\), hue shift and Gaussian noise, one per step. We optimize with AdamW (\(\beta_1 = 0.5\), \(\beta_2 = 0.999\), \(\epsilon = 10^{-6}\), weight decay \(10^{-5}\)) at a constant learning rate of \(10^{-4}\) and \(\tau=0\).

\begin{table}[!t]
    \centering
    \footnotesize
    \setlength{\tabcolsep}{2pt}
    \begin{tabular*}{\columnwidth}{@{\extracolsep{\fill}}lccc}
        \toprule
        method & CocoGlide & COVERAGE & CASIA v2.0 \\
        \midrule
        Ours w/ wm       & \pinkcell{\textbf{0.993/0.986}} & \pinkcell{\textbf{0.994/0.988}} & \pinkcell{0.989}/0.978 \\
        Ours w/o wm      & \textbf{0.993/0.986} & \textbf{0.994/0.988} & \textbf{0.989/0.979} \\
        \midrule
        GenPTW \cite{genptw2025} & 0.957/0.921 & 0.938/0.912 & 0.830/0.775 \\
        EditGuard \cite{zhang2024editguard} & 0.943/0.897 & 0.971/0.945 & 0.941/0.894 \\
        OmniGuard \cite{zhang2025omniguard} & 0.391/0.314 & 0.875/0.828 & 0.754/0.676 \\
        \midrule
        PSCC-Net \cite{liu2022pscc} & 0.254/0.176 & 0.335/0.229 & 0.297/0.234 \\
        ManTraNet \cite{wu2019mantra} & 0.387/0.259 & 0.260/0.163 & 0.172/0.109 \\
        FOCAL \cite{focal2023} & 0.562/0.477 & 0.728/0.672 & 0.894/0.848 \\
        \bottomrule
    \end{tabular*}
    \caption{Clean localization (F1/IoU) on CocoGlide \cite{glide2022,lin2014microsoft}, COVERAGE
    \cite{wen2016coveragea} and CASIA v2.0 \cite{dong2013casia} sets; ``w/o wm'' drops the payload. Bold: best of the
    column; shaded: best entry among the proactive methods.}
    \label{tab:exp_clean}
\end{table}

\begin{table*}[t]
    \centering
    \scriptsize
    \renewcommand{\arraystretch}{0.94}
    \setlength{\tabcolsep}{1.6pt}
    \begin{tabular*}{\textwidth}{@{\extracolsep{\fill}}l cccccc}
        \toprule
        method & FLUX.1 [schnell] & FLUX.1 [dev] & FLUX.1 Fill [dev] & Stable Diffusion 2 & Photoshop & mean \\
        \midrule
	Ours w/ wm   & \pinkcell{\textbf{0.991/0.974}} & \pinkcell{0.986/0.972} & \pinkcell{0.981/0.971} & \pinkcell{\textbf{0.972/0.967}} & \pinkcell{0.975/0.969} & \pinkcell{\textbf{0.981/0.971}} \\
	Ours w/o wm  & \textbf{0.991/0.974} & \textbf{0.986/0.973} & \textbf{0.981/0.972} & \textbf{0.972/0.967} & \textbf{0.975/0.970} & \textbf{0.981/0.971} \\
        \midrule
        EditGuard~\cite{zhang2024editguard}   & 0.929/0.877 & 0.932/0.881 & 0.927/0.875 & 0.927/0.875 & 0.930/0.880 & 0.929/0.878 \\
        OmniGuard~\cite{zhang2025omniguard}  & 0.850/0.786 & 0.785/0.715 & 0.783/0.707 & 0.900/0.848 & 0.943/0.900 & 0.852/0.791 \\
        GenPTW~\cite{genptw2025}             & 0.831/0.786 & 0.834/0.788 & 0.818/0.772 & 0.816/0.770 & 0.827/0.778 & 0.825/0.779 \\
        \midrule
        FOCAL~\cite{focal2023}               & 0.625/0.557 & 0.582/0.515 & 0.460/0.390 & 0.488/0.407 & 0.411/0.338 & 0.513/0.442 \\
        PSCC-Net~\cite{liu2022pscc}          & 0.345/0.265 & 0.251/0.189 & 0.301/0.229 & 0.232/0.170 & 0.413/0.324 & 0.309/0.235 \\
        ManTraNet~\cite{wu2019mantra}        & 0.285/0.189 & 0.223/0.144 & 0.298/0.203 & 0.314/0.210 & 0.602/0.478 & 0.345/0.245 \\
        \bottomrule
    \end{tabular*}
    \caption{Clean localization (F1/IoU) on the TGIF2 \cite{mareen2026tgif2}. ``w/o wm'' drops the payload, so the gap between the two VeriOrigin rows is its cost. Bold: best of the column; shaded: best entry among the proactive methods.}
    \label{tab:exp_quad}
\end{table*}

\begin{table*}[!t]
    \centering
    \scriptsize
    \setlength{\tabcolsep}{1.1pt}
    \renewcommand{\arraystretch}{0.94}
    \begin{tabular*}{\textwidth}{@{\extracolsep{\fill}}l c c ccc ccc ccc ccc ccc ccc}
        \toprule
        & & & \multicolumn{3}{c}{\textbf{JPEG}} & \multicolumn{3}{c}{\textbf{Gaussian noise}} & \multicolumn{3}{c}{\textbf{Gaussian blur}} & \multicolumn{3}{c}{\textbf{random crop}} & \multicolumn{3}{c}{\textbf{salt--pepper}} & \multicolumn{3}{c}{\textbf{resize}} \\
        \cmidrule(lr){4-6}\cmidrule(lr){7-9}\cmidrule(lr){10-12}\cmidrule(lr){13-15}\cmidrule(lr){16-18}\cmidrule(lr){19-21}
        method & res & tampered & 90 & 80 & 70 & 8 & 22 & 23 & 0.5 & 1.0 & 1.5 & 5\% & 15\% & 30\% & 0.5\% & 1\% & 2\% & 0.75 & 0.5 & 0.35 \\
        \midrule
        Ours w/ wm                          & 512 & \bestcell{1.000} & 0.988 & 0.966 & 0.923 & \bestcell{1.000} & \bestcell{0.999} & \bestcell{0.999} & \bestcell{1.000} & \bestcell{0.999} & \bestcell{0.999} & \bestcell{1.000} & 0.994 & 0.983 & 0.996 & 0.994 & 0.984 & \bestcell{1.000} & \bestcell{0.999} & \bestcell{0.999} \\
        \midrule
        OmniGuard~\cite{zhang2025omniguard} & 512 & 0.993 & 0.991 & 0.991 & \bestcell{0.990} & 0.991 & 0.971 & 0.968 & 0.993 & 0.993 & 0.992 & 0.989 & 0.956 & 0.852 & 0.976 & 0.953 & 0.909 & 0.993 & 0.993 & 0.992 \\
        EditGuard~\cite{zhang2024editguard}   & 512 & 0.998 & 0.862 & 0.837 & 0.820 & 0.964 & 0.725 & 0.712 & 0.994 & 0.468 & 0.209 & 0.997 & 0.991 & 0.972 & 0.999 & 0.999 & 0.998 & 0.989 & 0.179 & 0.286 \\
        GenPTW~\cite{genptw2025}             & 512 & 0.992 & 0.991 & 0.988 & 0.982 & 0.983 & 0.916 & 0.910 & 0.992 & 0.992 & 0.992 & 0.991 & 0.982 & 0.949 & 0.987 & 0.982 & 0.959 & 0.992 & 0.992 & 0.992 \\
        CRMark~\cite{chen2025crmark}         & 256 & \bestcell{1.000} & \bestcell{1.000} & \bestcell{0.999} & 0.984 & \bestcell{1.000} & 0.995 & 0.994 & \bestcell{1.000} & 0.968 & 0.414 & \bestcell{1.000} & \bestcell{1.000} & 0.793 & \bestcell{1.000} & \bestcell{1.000} & \bestcell{1.000} & \bestcell{1.000} & 0.561 & 0.429 \\
        TrustMark~\cite{trustmark}           & 256 & \bestcell{1.000} & 0.996 & 0.992 & 0.986 & 0.989 & 0.874 & 0.863 & \bestcell{1.000} & \bestcell{0.999} & \bestcell{0.999} & \bestcell{1.000} & 0.999 & 0.979 & 0.966 & 0.907 & 0.811 & \bestcell{1.000} & \bestcell{0.999} & \bestcell{0.999} \\
        InvisMark~\cite{xu2025invismark}     & 256 & 0.990 & 0.715 & 0.619 & 0.589 & 0.990 & 0.871 & 0.851 & 0.990 & 0.990 & 0.805 & 0.990 & 0.990 & \bestcell{0.990} & 0.990 & 0.990 & 0.990 & 0.990 & 0.990 & 0.576 \\
        \bottomrule
    \end{tabular*}
    \caption{Payload accuracy on TGIF2. Payload: VeriOrigin and OmniGuard 100 bits, CRMark, EditGuard and GenPTW 64, TrustMark and InvisMark 100. Bold: best of the column; columns group JPEG, noise, blur, crop, salt-pepper and resize.}
    \label{tab:exp_payload}
\vspace{-5pt}
\end{table*}

\begin{table}[!t]
    \centering
    \scriptsize
    \setlength{\tabcolsep}{6.2pt}
    \renewcommand{\arraystretch}{0.88}
    \resizebox{\columnwidth}{!}{%
    \begin{tabular}{l c c c c c}
        \toprule
        method & size & bits & PSNR & SSIM & LPIPS \\
        \midrule
        Ours w/ wm               & 512 & 100 & 31.95 & 0.841 & 0.0021 \\
        Ours w/o wm              & 512 & --  & 33.05 & 0.877 & 0.0008 \\
        \midrule
        CRMark \cite{chen2025crmark}   & 256 & 64  & 40.42 & 0.968 & 0.0005 \\
        TrustMark \cite{trustmark} & 256 & 100 & 41.64 & 0.985 & 0.0013 \\
        InvisMark \cite{xu2025invismark} & 256 & 100 & 42.79 & 0.993 & 0.0017 \\
        OmniGuard \cite{zhang2025omniguard} & 512 & 100 & 40.67 & 0.985 & 0.0036 \\
        GenPTW \cite{genptw2025}       & 512 & 64  & 37.58 & 0.969 & 0.0129 \\
        EditGuard \cite{zhang2024editguard}   & 512 & 64 & 36.28 & 0.933 & 0.0243 \\
        \bottomrule
    \end{tabular}}
    \caption{Visual quality on the TGIF2~\cite{mareen2026tgif2} covers at the native resolution and payload of each method.}
    \label{tab:exp_quality}
\end{table}

\subsection{Part I: Tamper Detection}
\label{sec:exp_local}

We evaluate our method against the proactive and the passive localisers on seven datasets. Because VeriOrigin couples a watermark with a localiser, we report two variants, ``w/ wm'' with the payload and ``w/o wm''
without it, to evaluate the influence of the watermark on tamper detection.

Table~\ref{tab:exp_clean} gives the clean results on CocoGlide, COVERAGE and CASIA v2.0; every baseline is re-run on full images under identical conditions. Against proactive methods, VeriOrigin reaches \(0.993\), \(0.994\) and \(0.989\) F1, ahead of EditGuard at \(0.943\), \(0.971\) and \(0.941\), GenPTW at \(0.957\), \(0.938\) and \(0.830\), and the embedding costs almost nothing in accuracy: the w/o wm rows stay within \(0.001\) of the w/ wm ones. OmniGuard is the one active baseline that behaves unevenly, \(0.875\) on COVERAGE but \(0.391\) on CocoGlide. Against passive methods, the margin is wider: FOCAL, the strongest of the three, reaches \(0.562\), \(0.728\) and \(0.894\), while PSCC-Net stays at \(0.254\), \(0.335\) and \(0.297\) and ManTraNet at \(0.387\), \(0.260\) and \(0.172\), since they look for tamper traces rather than the mark itself, which is what a proactive method changes.

Table~\ref{tab:exp_quad} breaks TGIF2 down by the inpainting model, FLUX.1 [schnell], [dev] and Fill [dev], Stable Diffusion 2 and Photoshop, \(100\) pairs each; Fig.~\ref{fig:exp_merged} shows four covers; every pair is scored on a seeded random \(512\times512\) crop of its native frame. Against proactive methods, VeriOrigin achieves a mean of \(0.981\) F1 and \(0.971\) IoU across five models, outperforming the strongest baseline, EditGuard, reaching \(0.929/0.878\). Dropping the payload yields virtually identical results, so the carrier costs almost no accuracy on clean channels. GenPTW follows at \(0.825/0.779\). Against passive localisers, the margin is even wider: FOCAL reaches \(0.513/0.442\), PSCC-Net \(0.309/0.235\), and ManTraNet \(0.345/0.245\).

In short, on clean channels, watermark embedding introduces minimal accuracy loss. Furthermore, VeriOrigin demonstrates highly competitive localization performance across datasets and editing models, consistently outperforming the active baselines, while passive detectors generally lag behind.

\begin{figure*}[t]
	\centering
	\includegraphics[width=\textwidth]{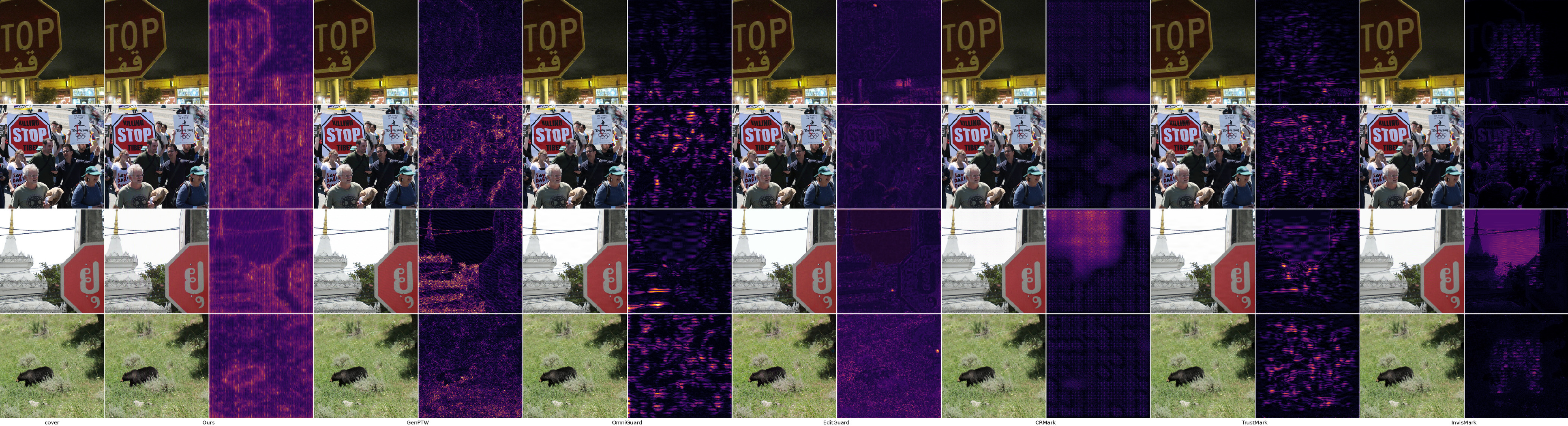}
	\caption{Stego (left) and $8\times$ residual (right) of each watermark on four random \(512\times512\) TGIF2 crops, at the native resolution of the method: VeriOrigin, GenPTW~\cite{genptw2025}, OmniGuard~\cite{zhang2025omniguard}, EditGuard~\cite{zhang2024editguard}, CRMark~\cite{chen2025crmark}, TrustMark~\cite{trustmark} and InvisMark~\cite{xu2025invismark}.}
    \label{fig:exp_wm}
\end{figure*}

\begin{figure*}[t]
	\centering
	\includegraphics[width=\textwidth]{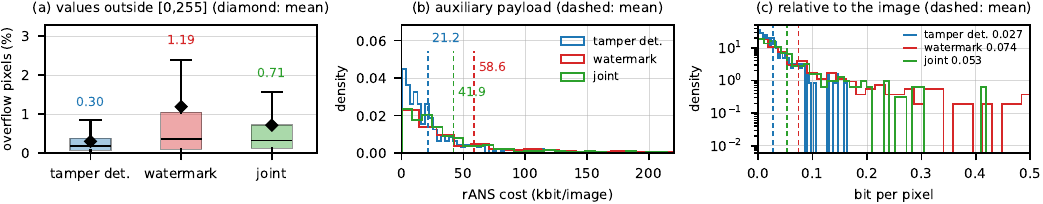}
	\caption{Auxiliary information of the embedding on TGIF2 for the three variants the pipeline touches: watermark branch only, tamper branch only, and the joint. (a) samples outside \([0,255]\), (b) rANS cost of the map, (c) that cost per pixel.}
	\label{fig:exp_reliable}
\end{figure*}

\begin{figure}[!t]
	\centering
	\includegraphics[width=\columnwidth]{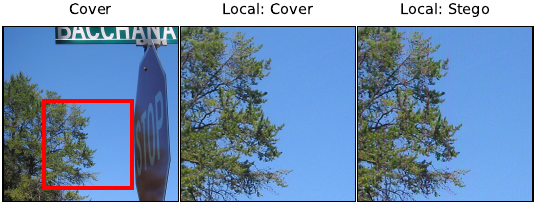}
	\caption{Global and local view of one embedding on a sample.}
	\label{fig:exp_mixed}
\end{figure}

\subsection{Part II: Payload Robustness}
\label{sec:exp_payload}

We use the same 500 TGIF2 crops for both halves of this part: visual quality against the cover by PSNR, SSIM and LPIPS, and payload robustness by the bit accuracy read back from the tampered image, over six distortion families at three strengths each, namely JPEG at quality 90, 80 and 70, Gaussian noise at \(\sigma=8\), 22 and 23, Gaussian blur at 0.5, 1.0 and 1.5, random crop at 5, 15 and 30\%, salt--pepper at 0.5, 1 and 2\% and resize at 0.75, 0.5 and 0.35, each method at its native resolution and on the same 500 crops.

\textbf{Visual quality.} Table~\ref{tab:exp_quality} evaluates the embedding distortion on 500 crops. The 256-resolution methods achieve the highest PSNR: InvisMark (42.79\,dB), TrustMark (41.64\,dB), and CRMark (40.42\,dB). Among the 512-resolution baselines, OmniGuard leads with 40.67\,dB, followed by GenPTW (37.58\,dB) and EditGuard (36.28\,dB), while our method achieves 31.95\,dB. While our method does not achieve the highest PSNR, it demonstrates strong competitiveness on LPIPS. VeriOrigin achieves 0.0021, which only trails behind three 256-resolution models (CRMark at 0.0005, TrustMark at 0.0013, and InvisMark at 0.0017) while being one to thirty times lower than all 512-resolution baselines, such as OmniGuard (0.0036), GenPTW (0.0129), and EditGuard (0.0243). This indicates that despite a relatively lower PSNR (31.95\,dB vs. 33.05\,dB without the payload), our watermark embedding achieves superior imperceptibility, a property that the LPIPS metric effectively captures.

Fig.~\ref{fig:exp_wm} compares the images of the different watermarking methods on four samples: the residual of VeriOrigin is highly content-adaptive. Fig.~\ref{fig:exp_mixed} provides a local zoom of an embedding result, showing the marked enlarged window and a side-by-side comparison of the cover and the embedded image within it. Distortions are predominantly concentrated in complex texture regions (such as foliage) and at the boundaries between objects and smooth regions, while the smooth sky remains virtually unaffected. This reveals that the embedding utilizes edge regions for localization by introducing localized distortions. This explains why VeriOrigin remains the least perceptible among the 512-resolution methods, despite its PSNR not being the highest.

\textbf{Robustness evaluation.} Table~\ref{tab:exp_payload} reports the payload accuracy after each distortion, with every method evaluated at its native resolution. Under compression, the reversible marks of CRMark, TrustMark, and GenPTW retain most of the payload down to quality \(70\). In contrast, VeriOrigin loses only a few points at the lowest quality, while InvisMark drops nearly a third of its payload already at quality \(90\). Under additive noise and blur, the order reverses: VeriOrigin remains at \(0.999\) for every strength, whereas the same reversible marks and EditGuard degrade rapidly as the perturbation grows, dropping to \(0.414\) and \(0.209\) for CRMark and EditGuard under the strongest blur. Cropping penalizes marks embedded in low-frequency content more heavily than VeriOrigin. Salt-and-pepper noise is the one family where CRMark remains virtually untouched and EditGuard nearly so, the signature of a spread-spectrum residual. Resizing is the weakest family for the reversible marks: at half resolution, VeriOrigin and TrustMark still return \(0.999\) and InvisMark \(0.990\), but CRMark drops to \(0.561\) and EditGuard to \(0.179\). Overall, these results demonstrate that VeriOrigin achieves highly competitive provenance performance across various distortions.

In short, VeriOrigin excels in both tamper localization and watermark robustness. It achieves the best localization results with minimal impact from the watermark embedding. Regarding robustness, it is the only method that maintains consistent accuracy across all distortion families, demonstrating highly competitive performance.

\begin{figure*}[!t]
	\centering
	\includegraphics[width=\textwidth]{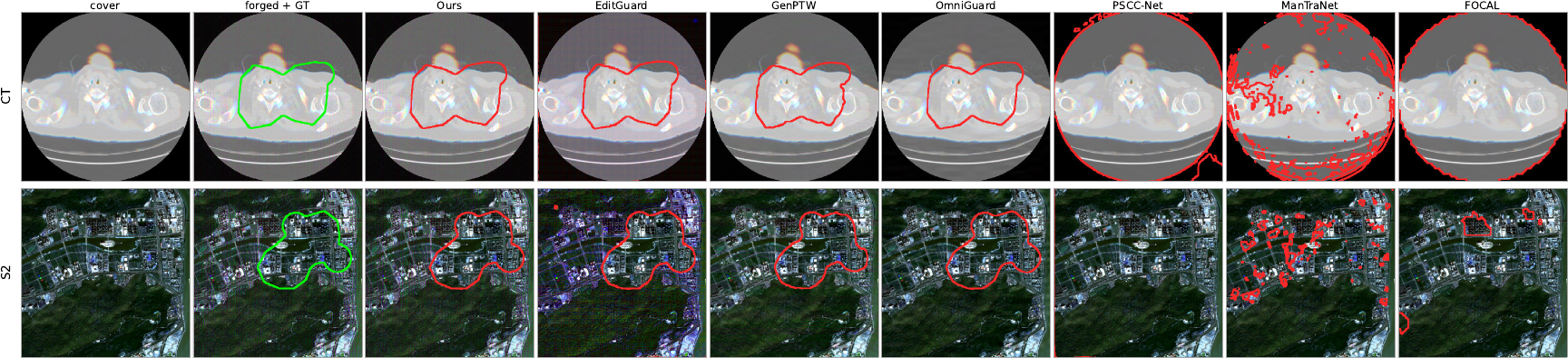}
	\caption{Localization on the two high bit-depth sets: a CT sample (top) and a Sentinel-2 chip (bottom).}
	\label{fig:s_hbd}
\vspace{-10pt}
\end{figure*}

\vspace{-5pt}
\subsection{Part III: Reversibility Analysis}
\label{sec:exp_reliable}
\label{sec:overflow}

Since integrity authentication requires the embedding process to be fully reversible, we retain auxiliary information for samples that fall outside the valid range \([0,255]\). We evaluate three variants on TGIF2: the payload-only branch, the tamper-only branch, and the final delivered joint image. For each image, we record the proportion of out-of-range samples and the rANS \cite{duda2013rans} coding cost of the map required for their restoration; Fig.~\ref{fig:exp_reliable} summarizes both metrics. The watermark branch alone incurs the highest overhead, with \(1.19\%\) of samples out of range and an rANS cost of 58.6\,kbit (0.0745 bit per pixel). The tamper carrier alone is the cheapest at \(0.30\%\), 21.2\,kbit, and 0.0270 bit per pixel. The delivered joint image lies between them at \(0.71\%\), 41.9\,kbit, and 0.0533 bit per pixel. This overhead is significantly below the capacity of RDH. Since many existing RDH methods can already achieve an embedding capacity of 1 bpp (bit per pixel), this ensures reversibility and further confirms the effectiveness of the overflow penalty loss during training.

\vspace{-10pt}
\subsection{Results on High Bit-Depth Data}\label{sec:s_bits}

This section evaluates the extension of our method to native-depth imagery. By applying the mapping of Section~\ref{sec:s_map}, we report its cost and accuracy on a 16-bit clinical CT set and a 16-bit Sentinel-2 L2A set.

\begin{table}[!t]
	\centering
	\footnotesize
	\setlength{\tabcolsep}{1.5pt}
	\begin{tabular*}{\columnwidth}{@{\extracolsep{\fill}}lccccc}
		\toprule
		\textbf{Set} & \textbf{Bits} & \textbf{PSNR} & \textbf{Ovf.}\textbf{(\%)} & \textbf{Aux}\textbf{(kbit)} & \textbf{Payload}\textbf{(\%)} \\
		\midrule
		LIDC-IDRI CT & 16 & 32.65 & 0.000 & 0.01 & 100.0 \\
		Sentinel-2 L2A & 16 & 34.07 & 0.000 & 0.01 & 100.0 \\
		\bottomrule
	\end{tabular*}
	\caption{Native-depth embedding of VeriOrigin on the two high bit-depth sets; PSNR is measured on the 8-bit render \(\lfloor(I-m_b)/s_b\rfloor\) that the network actually receives.}
	\label{tab:s_native}
\end{table}

\begin{table}[!t]
	\centering
	\footnotesize
	\setlength{\tabcolsep}{3pt}
	\begin{tabular*}{\columnwidth}{@{\extracolsep{\fill}}lcccc}
		\toprule
		\textbf{Method} & \multicolumn{2}{c}{\textbf{LIDC-IDRI CT}} & \multicolumn{2}{c}{\textbf{Sentinel-2 L2A}} \\
		\cmidrule(lr){2-3}\cmidrule(lr){4-5}
		& \textbf{F1} & \textbf{IoU} & \textbf{F1} & \textbf{IoU} \\
		\midrule
		\textbf{VeriOrigin (ours)} & \textbf{0.995} & \textbf{0.991} & \textbf{0.994} & \textbf{0.987} \\
		\midrule
		EditGuard~\cite{zhang2024editguard} & 0.980 & 0.961 & 0.938 & 0.898 \\
		GenPTW~\cite{genptw2025} & 0.986 & 0.972 & 0.985 & 0.971 \\
		OmniGuard~\cite{zhang2025omniguard} & 0.985 & 0.971 & 0.957 & 0.932 \\
		PSCC-Net~\cite{liu2022pscc} & 0.313 & 0.189 & 0.288 & 0.174 \\
		ManTraNet~\cite{wu2019mantra} & 0.063 & 0.033 & 0.043 & 0.023 \\
		FOCAL~\cite{focal2023} & 0.002 & 0.001 & 0.072 & 0.041 \\
		\bottomrule
	\end{tabular*}
	\caption{Localization on the two high bit-depth sets, all \(400\) CT samples and \(360\) chips.}
	\label{tab:s_native_loc}
\end{table}

\textbf{Clinical and remote-sensing images.} The high bit-depth half uses two public sources: the CT series of three LIDC-IDRI subjects~\cite{armato2011lidc}, \(400\) slices in total, whose slices are \(512\times512\) DICOM with BitsStored \(16\) and are shifted from the signed into the unsigned range by \(2^{15}\) before embedding and shifted back after recovery, and two Sentinel-2 L2A acquisitions~\cite{sentinel2} of \(1129\times1031\) pixels with six \(16\)-bit bands, where bands \(1\)--\(3\) and \(4\)--\(6\) form two three-channel groups and \(512\times512\) windows with stride \(64\) give \(360\) chips. Each sample stacks three consecutive slices into the three channels, giving \(400\) CT samples from the three series, and all \(360\) Sentinel-2 chips are evaluated. The evaluation is zero-shot. \Cref{tab:s_native} reports \(32.65\) dB on CT and \(34.07\) dB on Sentinel-2, measured on that 8-bit render; no sample leaves the range on either set, so the auxiliary stream that remains is \(0.008\) kbit under 16-bit embedding. Every one of the \(760\) images returns its bits with the cover restored bit for bit.

\begin{figure*}[!t]
	\centering
	\includegraphics[width=\textwidth]{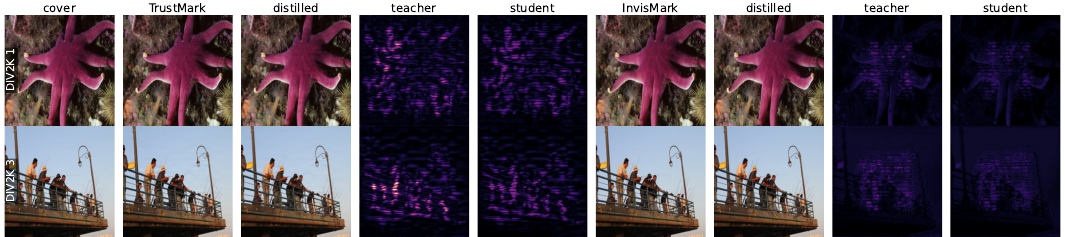}
	\caption{Visualization comparison of teacher and student on two DIV2K covers.}
	\label{fig:s_wmdistill}
\end{figure*}

\begin{table*}[!t]
	\centering
	\scriptsize
	\setlength{\tabcolsep}{1.8pt}
	\renewcommand{\arraystretch}{0.92}
	\begin{tabular*}{\textwidth}{@{\extracolsep{\fill}}l c ccc ccc ccc ccc ccc ccc ccc}
		\toprule
		& & \multicolumn{3}{c}{\textbf{JPEG}} & \multicolumn{3}{c}{\textbf{Gaussian noise}} & \multicolumn{3}{c}{\textbf{Gaussian blur}} & \multicolumn{3}{c}{\textbf{random crop}} & \multicolumn{3}{c}{\textbf{color}} & \multicolumn{3}{c}{\textbf{salt--pepper}} & \multicolumn{3}{c}{\textbf{resize}} \\
		\cmidrule(lr){3-5}\cmidrule(lr){6-8}\cmidrule(lr){9-11}\cmidrule(lr){12-14}\cmidrule(lr){15-17}\cmidrule(lr){18-20}\cmidrule(lr){21-23}
		model & clean & 90 & 80 & 70 & 8 & 22 & 23 & 0.5 & 1.0 & 1.5 & 5\% & 15\% & 30\% & 0.2 & 0.4 & 0.6 & 0.5\% & 1\% & 2\% & 0.75 & 0.5 & 0.35 \\
		\midrule
		TrustMark (teacher) & 99.9 & 99.6 & 99.2 & 98.9 & 99.2 & 91.4 & 91.0 & 99.9 & 99.9 & 99.8 & 99.9 & 99.9 & 92.5 & 99.6 & 95.3 & 89.9 & 99.8 & 97.7 & 94.0 & 99.9 & 99.9 & 99.9 \\
		TrustMark (distilled) & 98.6 & 96.9 & 95.9 & 93.6 & 94.9 & 85.3 & 84.2 & 98.6 & 97.9 & 97.6 & 98.5 & 97.2 & 89.2 & 95.9 & 90.1 & 83.3 & 96.8 & 93.9 & 84.1 & 98.2 & 97.6 & 97.4 \\
		InvisMark (teacher) & 95.5 & 65.4 & 59.3 & 57.2 & 93.4 & 74.1 & 71.8 & 95.5 & 95.5 & 62.8 & 95.5 & 95.5 & 95.1 & 94.8 & 93.4 & 91.2 & 95.4 & 95.6 & 95.2 & 95.5 & 95.4 & 53.1 \\
		InvisMark (distilled) & 95.1 & 59.6 & 57.0 & 56.4 & 92.8 & 67.8 & 64.8 & 95.1 & 95.0 & 59.1 & 95.1 & 95.1 & 95.0 & 94.0 & 92.3 & 89.9 & 95.1 & 95.1 & 94.9 & 94.9 & 94.9 & 52.9 \\
		\bottomrule
	\end{tabular*}
	\caption{Payload accuracy (\%) of the teachers and their distilled students.}
	\label{tab:s_wm_rob}
\end{table*}

\begin{table}[!t]
	\centering
	\scriptsize
	\setlength{\tabcolsep}{2.4pt}
	\renewcommand{\arraystretch}{0.96}
	\begin{tabular*}{\columnwidth}{@{\extracolsep{\fill}}lcccccc}
		\toprule
		\textbf{Model} & \textbf{PSNR} & \textbf{SSIM} & \textbf{LPIPS} & \textbf{Pay.} & \textbf{Ovf.} & \textbf{Aux}(bit) \\
		\midrule
		\multicolumn{7}{@{}l}{\emph{DIV2K (200)}}\\
		TrustMark (teacher) & 40.39 & 0.9859 & 0.0011 & 99.90 & -- & -- \\
		\quad distilled & 41.82 & 0.9888 & 0.0009 & 98.43 & 0.010 & 303 \\
		InvisMark (teacher) & 43.20 & 0.9956 & 0.0008 & 95.89 & -- & -- \\
		\quad distilled & 44.55 & 0.9970 & 0.0006 & 95.61 & 0.010 & 293 \\
		\midrule
		\multicolumn{7}{@{}l}{\emph{TGIF2 (200)}}\\
		TrustMark (teacher) & 41.39 & 0.9847 & 0.0012 & 99.94 & -- & -- \\
		\quad distilled & 42.47 & 0.9873 & 0.0011 & 99.50 & 0.004 & 143 \\
		InvisMark (teacher) & 42.83 & 0.9941 & 0.0012 & 95.50 & -- & -- \\
		\quad distilled & 43.85 & 0.9957 & 0.0008 & 95.21 & 0.004 & 141 \\
		\bottomrule
	\end{tabular*}
	\caption{Teachers and distilled students.}
	\label{tab:s_distill}
\end{table}

Since neither set ships a tamper mask, each sample is forged by drawing one irregular mask over \(6\)--\(26\%\) of the frame and pasting back the original content under it after a mild perspective warp, so the edit displaces the content inside the mask instead of restoring a clean copy. As reported in Table~\ref{tab:s_native_loc}, on the 8-bit quantized image, our localiser reaches \(0.995\) F1 on CT and \(0.994\) on Sentinel-2. Among the proactive baselines GenPTW is the closest at \(0.986\) and \(0.985\), followed by OmniGuard at \(0.985\) and \(0.957\) and EditGuard at \(0.980\) and \(0.938\), and on Sentinel-2 it is overtaken by GenPTW. The passive detectors trail far behind, at \(0.313\) and \(0.288\) for PSCC-Net, \(0.063\) and \(0.043\) for ManTraNet and \(0.002\) and \(0.072\) for FOCAL. This gap exists because the edit displaces the content inside the mask, and the watermark branch of VeriOrigin models exactly that displacement. \Cref{fig:s_hbd} visualizes the two rows of results.

Table~\ref{tab:s_native} reports the number of overflow pixels and the auxiliary bitstream length. We embed into all \(400\) CT samples and \(360\) chips. Because CT images are mostly black with a concentrated grayscale distribution, the residuals are predominantly negative and highly prone to downward overflow. Therefore, applying a global integer shift \(t\) in native units before embedding eliminates most of this downward overflow, introducing only a small constant offset. With the per-image scale and this shift no sample of either set leaves \([0, 2^{16}-1]\), so the auxiliary stream is essentially empty at \(0.008\) kbit and the 8-bit delivered image stays at \(32.65\)\,dB on CT and \(34.07\)\,dB on Sentinel-2. This confirms that the lossless mapping carries the native depth in the residual rather than re-encoding the frame: all \(100\) bits are returned, and the arithmetic step of \Cref{lem:lossless_mapping} holds on all \(760\) images. On the 8-bit render the localiser reaches \(0.995\) and \(0.994\) F1.

In short, the proposed lossless mapping proves highly effective for high-bit-depth data. The per-image scale keeps the render within range and the shift \(t\) removes the remaining downward overflow, so both sets run with an essentially empty side stream, watermarks are reliably recovered, and the localiser stays above \(0.99\) F1 on both.

\vspace{-10pt}
\subsection{Distilled Watermark Results}
\label{sec:s_distill}
Beyond extending to higher bit-depth imagery, we further distill our proposed invertible encoder, XONet, into existing watermarking methods to endow them with invertibility. We selected two representative methods, TrustMark~\cite{trustmark} and InvisMark~\cite{xu2025invismark}, and conducted detailed distillation experiments on them, as described below.

\subsubsection{Distilling a Robust Watermark into XOnet}
\label{sec:s_drecipe}

Both TrustMark and InvisMark operate at \(256\times256\) with a 100-bit payload. We build the distillation set by querying the frozen teachers on DIV2K covers, recording tuples of cover, watermarked image. The student is our XOnet, trained on these tuples with the loss in \Cref{eq:s_distill}, a learning rate of \(10^{-4}\), and batch sizes of 96 for TrustMark and 112 for InvisMark. At test time the payload is extracted by the teacher's frozen decoder. Importantly, we fix \(\tau=1\) for XONet throughout distillation, so the reported numbers match a deployed model.
\begin{equation}
	\mathcal{L} = \lambda_{\text{mse}}\mathcal{L}_{\text{mse}}(\mathbf{S}_s, \mathbf{S}_t) + \lambda_{\text{pen}}(e) \mathcal{L}_{\text{pen}}(\mathbf{S}_s),
	\label{eq:s_distill}
\end{equation}
where \(\lambda_{\text{mse}} = 1\), \(\mathbf{S}_s\) and \(\mathbf{S}_t\) denote the student and teacher stego images, and \(\mathcal{L}_{\text{pen}}\) is the overflow penalty measuring the mean squared distance of samples outside \([0, 1]\). The parameters for dynamic loss adjustment are \(e_0=0, \Delta=5\). We train the model using the aforementioned loss function until full convergence is achieved.

\begin{figure*}[!t]
	\centering
	\includegraphics[width=\textwidth]{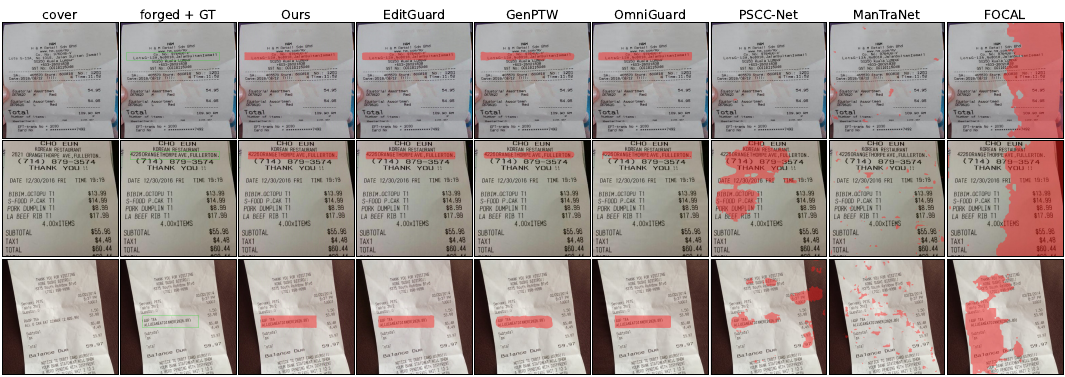}
	\caption{Document tampering on three AIForge-Doc samples.}
	\label{fig:s_docs}
\end{figure*}

\begin{table*}[!t]
	\centering
	\scriptsize
	\setlength{\tabcolsep}{1.6pt}
	\renewcommand{\arraystretch}{0.96}
	\begin{tabular*}{\textwidth}{@{\extracolsep{\fill}}l c ccc ccc ccc ccc ccc ccc}
		\toprule
		& & \multicolumn{3}{c}{\textbf{JPEG}} & \multicolumn{3}{c}{\textbf{Gaussian noise}} & \multicolumn{3}{c}{\textbf{Gaussian blur}} & \multicolumn{3}{c}{\textbf{random crop}} & \multicolumn{3}{c}{\textbf{salt--pepper}} & \multicolumn{3}{c}{\textbf{resize}} \\
		\cmidrule(lr){3-5}\cmidrule(lr){6-8}\cmidrule(lr){9-11}\cmidrule(lr){12-14}\cmidrule(lr){15-17}\cmidrule(lr){18-20}
		method & clean & 90 & 80 & 70 & 8 & 22 & 23 & 0.5 & 1.0 & 1.5 & 5\% & 15\% & 30\% & 0.5\% & 1\% & 2\% & 0.75 & 0.5 & 0.35 \\
		\midrule
		\textbf{VeriOrigin (ours)} & 1.000 & 0.984 & 0.952 & 0.887 & 1.000 & 1.000 & 1.000 & 1.000 & 1.000 & 1.000 & 1.000 & 0.996 & 0.990 & 1.000 & 1.000 & 0.998 & 1.000 & 1.000 & 1.000 \\
		\midrule
		EditGuard~\cite{zhang2024editguard} & 0.971 & 0.920 & 0.891 & 0.869 & 0.934 & 0.699 & 0.689 & 0.982 & 0.468 & 0.202 & 0.969 & 0.967 & 0.944 & 0.977 & 0.980 & 0.985 & 0.999 & 0.195 & 0.253 \\
		GenPTW~\cite{genptw2025} & 1.000 & 0.998 & 0.989 & 0.970 & 0.979 & 0.877 & 0.869 & 1.000 & 1.000 & 0.999 & 0.999 & 0.996 & 0.975 & 1.000 & 0.999 & 0.998 & 1.000 & 1.000 & 1.000 \\
		OmniGuard~\cite{zhang2025omniguard} & 1.000 & 1.000 & 1.000 & 1.000 & 0.999 & 0.984 & 0.982 & 1.000 & 1.000 & 1.000 & 1.000 & 0.989 & 0.895 & 1.000 & 0.998 & 0.989 & 1.000 & 1.000 & 1.000 \\
		CRMark~\cite{chen2025crmark} & 1.000 & 1.000 & 0.999 & 0.992 & 1.000 & 0.999 & 0.998 & 1.000 & 0.996 & 0.404 & 1.000 & 1.000 & 0.796 & 1.000 & 1.000 & 1.000 & 1.000 & 0.575 & 0.417 \\
		TrustMark~\cite{trustmark} & 1.000 & 0.994 & 0.982 & 0.976 & 0.978 & 0.773 & 0.757 & 1.000 & 1.000 & 1.000 & 1.000 & 1.000 & 0.983 & 0.997 & 0.978 & 0.943 & 1.000 & 1.000 & 0.999 \\
		InvisMark~\cite{xu2025invismark} & 0.990 & 0.686 & 0.609 & 0.563 & 0.990 & 0.813 & 0.784 & 0.990 & 0.990 & 0.926 & 0.990 & 0.990 & 0.990 & 0.990 & 0.990 & 0.990 & 0.990 & 0.990 & 0.562 \\
		\bottomrule
	\end{tabular*}
	\caption{Payload accuracy on all \(613\) AIForge-Doc samples.}
	\label{tab:s_doc_pay}

\vspace{-5pt}
\end{table*}

\begin{table}[!t]
	\centering
	\footnotesize
	\setlength{\tabcolsep}{3pt}
	\renewcommand{\arraystretch}{0.92}
	\begin{tabular*}{\columnwidth}{@{\extracolsep{\fill}}lcccccc}
		\toprule
		\textbf{Method} & \textbf{Bits} & \textbf{PSNR} & \textbf{SSIM} & \textbf{F1} & \textbf{IoU} \\
		\midrule
		VeriOrigin (ours) & 100 & 32.37 & 0.809 & \textbf{0.937} & \textbf{0.908} \\
		\midrule
		EditGuard~\cite{zhang2024editguard} & 64 & 35.96 & 0.908 & 0.807 & 0.747 \\
		OmniGuard~\cite{zhang2025omniguard} & 100 & 42.86 & 0.988 & 0.451 & 0.403 \\
		GenPTW~\cite{genptw2025} & 64 & 38.75 & 0.976 & 0.136 & 0.112 \\
		\midrule
		FOCAL~\cite{focal2023} & -- & -- & -- & 0.044 & 0.029 \\
		PSCC-Net~\cite{liu2022pscc} & -- & -- & -- & 0.030 & 0.017 \\
		ManTraNet~\cite{wu2019mantra} & -- & -- & -- & 0.015 & 0.008 \\
		\bottomrule
	\end{tabular*}
	\caption{Clean localization results on AIForge-Doc samples.}
	\label{tab:s_doc_loc}
\end{table}

\subsubsection{Experimental Results on Distilling Task}
\label{sec:s_dres}

\Cref{tab:s_distill,tab:s_wm_rob} present the quantitative results, and \Cref{fig:s_wmdistill} visualizes four sample cover images from DIV2K. On clean inputs, the student models recover \(98.43\%\) and \(95.61\%\) of the message on DIV2K, and \(99.50\%\) and \(95.21\%\) on TGIF2, compared with \(99.94\%\) and \(95.50\%\) for their teachers on the same set, and \(99.90\%\) and \(95.89\%\) on DIV2K. Notably, the students achieve slightly higher fidelity: \(41.82\) vs.\ \(40.39\) dB and \(44.55\) vs.\ \(43.20\) dB on DIV2K, and \(42.47\) vs.\ \(41.39\) dB and \(43.85\) vs.\ \(42.83\) dB on TGIF2. SSIM and LPIPS exhibit a consistent trend. The reversibility absent in the teachers incurs no practical overhead: across the \(400\) cover images, the students place at most \(0.010\%\) of samples out of the valid range, and the side stream that codes those samples is only \(141\)--\(303\) bits per image, far below the RDH capacity.

\Cref{tab:s_wm_rob} reports the full grid across seven distortion families and three severity levels. Averaged over the twenty-one distorted conditions, the TrustMark and InvisMark students successfully retain the vast majority of the teacher's extraction capability, with only a \(3.8\) and \(1.6\) point drop, respectively. The students perform best under blur and resize, where the gap stays within \(2.5\) points, while the strongest blur costs the InvisMark student \(3.7\) and the harshest crop the TrustMark student \(3.3\), and the InvisMark student is particularly robust to salt-pepper noise, differing by only \(0.4\) points. The worst cases occur under noise and JPEG. However, these represent successful knowledge transfer rather than newly introduced failure modes: on JPEG, for instance, the teacher itself drops to \(59.3\%\) and the student to \(57.0\%\), faithfully reproducing the degradation trend of its teacher. 

This indicates that XONet can indeed learn the teacher's watermark embedding through distillation. Overall, distillation successfully makes an existing robust mark reversible, keeps its fidelity within \(1.5\)\,dB, and preserves \(82.7\)--\(93.7\%\) of its payload accuracy on average. The students merely inherit the teacher's weak spots rather than developing new ones. We believe that incorporating decoder distillation would further enhance the extraction performance.

\subsection{Additional Experimental Results}
\label{sec:s_extra}

This section presents two experiments: a document benchmark for forgery localization and provenance, and an evaluation of embedding strength. We demonstrate that reducing strength shortens the auxiliary bitstream to improve the embedding success rate, and analyze the trade-off among strength, distortion, and performance.

\subsubsection{Document Images}
\label{sec:s_docs}

Document forgery poses severe security risks, since subtle manual tampering on bright document pages is hard to spot and threatens the authenticity of archival and evidentiary materials. We evaluate on the full testing split of AIForge-Doc~\cite{aiforgedoc}, comprising \(613\) triples of a pristine page, an AI-forged page and the field mask; each page is resized to \(512\times512\) and the shipped mask is kept, with fields covering \(0.02\) to \(5.3\%\) of a page (median \(0.8\%\)). The pristine page is watermarked, the forged field is spliced back through the mask, and the detector reads the delivered image; \Cref{fig:s_docs} shows three of the 613 triples. \Cref{tab:s_doc_loc} reports the clean case: VeriOrigin achieves \(0.937\) F1 and \(0.908\) IoU, significantly outperforming EditGuard (\(0.807\)/\(0.747\)) by a margin of \(13.0\) points. OmniGuard follows at \(0.451\) and GenPTW at \(0.136\), while the passive detectors stay below \(0.05\), an order of magnitude behind ours.

The embedded mark incurs \(32.37\) dB and \(0.809\) SSIM on the delivered page, compared to \(35.96\) dB and \(0.908\) SSIM for EditGuard, and \(38.75\) dB and \(42.86\) dB for GenPTW and OmniGuard, maintaining a similar trade-off as on natural images. \Cref{tab:s_doc_pay} reports the payload after every distortion. It stays at \(1.000\) under noise, blur and resize, and \(0.996\) under a \(15\%\) crop. Under JPEG compression at qualities \(90\), \(80\), and \(70\), it maintains accuracies of \(0.984\), \(0.952\), and \(0.887\), respectively. In contrast, EditGuard degrades rapidly, dropping to \(0.869\) at JPEG quality \(70\), \(0.699\) under noise, \(0.468\) under blur, and merely \(0.195\) at half resolution.

In short, VeriOrigin demonstrates competitive performance in document forgery localization, generally outperforming both active and passive baselines. Furthermore, it maintains a favorable balance among visual fidelity, payload capacity, and robustness. Even under severe distortions, the embedded payload remains largely intact, securing the provenance. This suggests that the framework is practical for archival and evidentiary use, since a page is watermarked once and re-analysed years later without losing its provenance.

\begin{figure*}[!t]
	\centering
	\includegraphics[width=\textwidth]{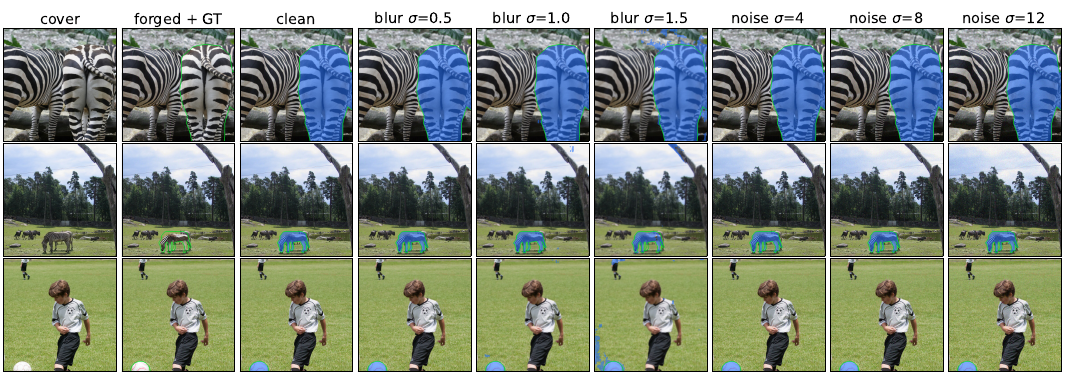}
	\caption{Predicted tamper masks under the three severity levels of blur and noise, three randomly drawn TGIF2 samples.}
	\label{fig:s_dist}
\end{figure*}

\begin{table*}[!t]
	\centering
	\footnotesize
	\setlength{\tabcolsep}{3pt}
	\renewcommand{\arraystretch}{0.98}
	\begin{tabular*}{\textwidth}{@{\extracolsep{\fill}}l ccc cc c ccc ccc}
		\toprule
		& \multicolumn{3}{c}{\textbf{Fidelity}} & \multicolumn{2}{c}{\textbf{Side map}} & \textbf{clean} & \multicolumn{3}{c}{\textbf{Gaussian blur (\(\sigma\))}} & \multicolumn{3}{c}{\textbf{Gaussian noise (\(\sigma\))}} \\
		\cmidrule(lr){2-4}\cmidrule(lr){5-6}\cmidrule(lr){7-7}\cmidrule(lr){8-10}\cmidrule(lr){11-13}
		\(\alpha_O=\alpha_X\) & PSNR & SSIM & LPIPS & ovf.(\%) & aux (kbit) & F1/IoU & 0.5 & 1.0 & 1.5 & 4 & 8 & 12 \\
		\midrule
		1.00 & 31.95 & 0.8410 & 0.0021 & 0.710 & 41.90 & 0.986/0.972 & 0.968/0.939 & 0.857/0.786 & 0.421/0.317 & 0.979/0.960 & 0.967/0.937 & 0.959/0.925 \\
		0.98 & 32.13 & 0.8414 & 0.0021 & 0.700 & 40.91 & 0.986/0.972 & 0.968/0.939 & 0.857/0.786 & 0.420/0.317 & 0.979/0.960 & 0.967/0.937 & 0.959/0.925 \\
		0.96 & 32.31 & 0.8443 & 0.0021 & 0.690 & 39.93 & 0.985/0.972 & 0.967/0.938 & 0.861/0.790 & 0.415/0.313 & 0.979/0.959 & 0.966/0.936 & 0.958/0.923 \\
		0.94 & 32.49 & 0.8507 & 0.0020 & 0.681 & 38.96 & 0.985/0.971 & 0.966/0.937 & 0.864/0.793 & 0.402/0.301 & 0.978/0.958 & 0.964/0.934 & 0.956/0.919 \\
		0.92 & 32.68 & 0.8571 & 0.0019 & 0.672 & 37.98 & 0.985/0.970 & 0.965/0.935 & 0.862/0.790 & 0.389/0.288 & 0.978/0.958 & 0.963/0.932 & 0.954/0.917 \\
		0.90 & 32.87 & 0.8640 & 0.0018 & 0.662 & 37.00 & 0.984/0.970 & 0.965/0.935 & 0.861/0.789 & 0.361/0.263 & 0.977/0.957 & 0.963/0.930 & 0.953/0.914 \\
		\bottomrule
	\end{tabular*}
	\caption{Embedding strength in \([0.90,1.00]\) with \(\alpha_O=\alpha_X=\alpha\), on \(500\) TGIF2 samples.}
	\label{tab:s_alpha}
\end{table*}

\begin{table*}[!t]
	\centering
	\scriptsize
	\setlength{\tabcolsep}{1.1pt}
	\renewcommand{\arraystretch}{0.94}
	\begin{tabular*}{\textwidth}{@{\extracolsep{\fill}}l c ccc ccc ccc ccc ccc ccc}
		\toprule
		& & \multicolumn{3}{c}{\textbf{JPEG}} & \multicolumn{3}{c}{\textbf{Gaussian noise}} & \multicolumn{3}{c}{\textbf{Gaussian blur}} & \multicolumn{3}{c}{\textbf{random crop}} & \multicolumn{3}{c}{\textbf{salt--pepper}} & \multicolumn{3}{c}{\textbf{resize}} \\
		\cmidrule(lr){3-5}\cmidrule(lr){6-8}\cmidrule(lr){9-11}\cmidrule(lr){12-14}\cmidrule(lr){15-17}\cmidrule(lr){18-20}
		\(\alpha_O=\alpha_X\) & clean & 90 & 80 & 70 & 8 & 22 & 23 & 0.5 & 1.0 & 1.5 & 5\% & 15\% & 30\% & 0.5\% & 1\% & 2\% & 0.75 & 0.5 & 0.35 \\
		\midrule
		1.00 & 1.000 & 0.988 & 0.966 & 0.923 & 1.000 & 0.999 & 0.999 & 1.000 & 0.999 & 0.999 & 1.000 & 0.994 & 0.983 & 0.996 & 0.994 & 0.984 & 1.000 & 0.999 & 0.999 \\
		0.98 & 1.000 & 0.988 & 0.966 & 0.924 & 1.000 & 0.999 & 0.999 & 1.000 & 0.999 & 0.999 & 1.000 & 0.994 & 0.983 & 0.997 & 0.996 & 0.995 & 0.999 & 0.999 & 0.999 \\
		0.96 & 1.000 & 0.988 & 0.963 & 0.918 & 0.999 & 0.999 & 0.999 & 1.000 & 0.999 & 0.999 & 1.000 & 0.994 & 0.982 & 0.997 & 0.996 & 0.994 & 0.999 & 0.999 & 0.999 \\
		0.94 & 1.000 & 0.986 & 0.960 & 0.910 & 0.999 & 0.999 & 0.999 & 1.000 & 0.999 & 0.999 & 1.000 & 0.994 & 0.982 & 0.997 & 0.996 & 0.994 & 0.999 & 0.999 & 0.999 \\
		0.92 & 1.000 & 0.985 & 0.956 & 0.903 & 0.999 & 0.999 & 0.998 & 1.000 & 0.999 & 0.999 & 1.000 & 0.994 & 0.982 & 0.997 & 0.995 & 0.994 & 0.999 & 0.999 & 0.998 \\
		0.90 & 1.000 & 0.984 & 0.951 & 0.897 & 0.999 & 0.998 & 0.998 & 1.000 & 0.999 & 0.999 & 1.000 & 0.994 & 0.982 & 0.997 & 0.995 & 0.993 & 0.999 & 0.999 & 0.998 \\
		\bottomrule
	\end{tabular*}
	\caption{Watermark robustness of the same sweep on TGIF2.}
	\label{tab:s_alpha_pay}
\vspace{-10pt}
\end{table*}

\subsubsection{Embedding Strength}
\label{sec:s_alpha}
Reducing the embedding strength is primarily motivated by shrinking the auxiliary bitstream and mitigating pixel overflow, which jointly increases the embedding success rate by lowering the capacity burden on the RDH channel. To quantify this trade-off, we set both branch strengths to the same value, \(\alpha_O=\alpha_X=\alpha\), which scales the residual written by both branches. Because it multiplies the delivered residual, \(\alpha\) can be adjusted at inference without retraining. We sweep \(\alpha \in [0.90, 1.00]\) on the \(500\) TGIF2 crops, each with its forged field spliced through the mask as in the document protocol, so the clean column differs from the corpus pairs of \Cref{tab:exp_quad}; \Cref{tab:s_alpha} reports the resulting performance. As \(\alpha\) decreases from \(1.00\) to \(0.90\), the overflow rate drops from \(0.710\%\) to \(0.662\%\) and the side map shrinks from \(41.9\) to \(37.0\) kbit. This reduction in auxiliary information significantly eases the embedding process and enhances the success rate. Concurrently, fidelity scales almost linearly with \(\alpha\): the delivered image gains \(0.92\)\,dB, SSIM rises from \(0.841\) to \(0.864\), and LPIPS falls from \(0.0021\) to \(0.0018\).

Crucially, reducing the embedding strength exerts only a minor impact on the overall localization performance. The clean F1 moves slightly from \(0.986\) to \(0.984\) (IoU \(0.972\) to \(0.970\)), and the mild blur and noise columns stay well within half a point. Only the hardest distortion columns experience a small drop, with blur \(\sigma{=}1.5\) falling from \(0.421\) to \(0.361\) and noise \(\sigma{=}12\) from \(0.959\) to \(0.953\). \Cref{fig:s_dist} further visualizes the predicted tamper masks under varying distortions, confirming that the localiser still robustly identifies forged regions even under such degradation.

\textbf{Watermark robustness.} \Cref{tab:s_alpha_pay} reports the payload after each distortion of the same sweep. On clean images, accuracy remains perfectly stable at \(1.000\). Across most distortions (noise, blur, cropping), accuracy fluctuates by at most \(0.005\). The only noticeable degradation occurs under strong JPEG compression, where accuracy drops by merely \(1.5\) points at quality \(80\) and \(2.6\) points at quality \(70\). This asymmetry makes \(\alpha\) a practical inference-time knob: setting \(\alpha=0.98\) keeps the localiser within \(0.001\) and the payload within \(0.011\) of their \(\alpha=1.00\) values, while recovering most of the fidelity gains and further reducing the side map.

In short, lowering \(\alpha\) effectively shrinks the auxiliary bitstream to guarantee successfully embedding, while simultaneously improving fidelity with only a slight compromise on localization and payload extraction.

\vspace{-5pt}
\section{Conclusion}
\label{sec:conclusion}

We presented VeriOrigin, a comprehensive and versatile framework that seamlessly unifies provenance, tamper localization, and lossless cover recovery. At its core, XOnet is a strictly invertible CNN-based encoder that hosts standard CNN watermarks, enabling the payload and the tamper residual to coexist in a single image with a remarkably low LPIPS. By decoupling into two orthogonal branches, VeriOrigin supports independent training and upgrading of the payload and localization stages without requiring any retraining. Furthermore, it natively supports high-bit-depth archives, such as 16-bit CT and satellite imagery, broadening its practical applicability. Moreover, XOnet can distill existing robust watermarking models at deployment time, preserving their original robustness while endowing them with strict structural invertibility. Extensive experiments across seven distinct datasets demonstrate that VeriOrigin not only guarantees perfect original cover recovery in lossless channels for integrity authentication, but also achieves robust watermark provenance and highly competitive tamper localization capabilities.
\bibliographystyle{IEEEtran}
\bibliography{refs_all}

\end{document}